\documentclass[twoside,11pt]{article}

\usepackage{blindtext}

 \usepackage[abbrvbib, preprint]{jmlr2e}

\usepackage{jmlr2e}

\usepackage[T1]{fontenc}
\usepackage{subcaption}
\usepackage{amsmath}
\usepackage{algorithm}
\usepackage{algpseudocode}
\usepackage{float}
\usepackage{booktabs}
\usepackage{tabularx}
\usepackage{lastpage}

\jmlrheading{00}{2025}{1-\pageref{LastPage}}{1/21; Revised 5/22}{9/22}{21-0000}{Huiqi Zhang and Fang Xie}

\ShortHeadings{AdaDPIGU: Private SGD with Adaptive Clipping}{Zhang and Xie}

\firstpageno{1}

\begin{document}

\title{AdaDPIGU: Differentially Private SGD with Adaptive Clipping and Importance-Based Gradient Updates for Deep Neural Networks}

\author{\name Huiqi Zhang \email zhanghq\_0111@163.com \\
       \addr Guangdong Provincial Key Laboratory of IRADS\\
             Beijing Normal-Hong Kong Baptist University\\
             Zhuhai 519087, China
       \AND
       \name Fang Xie\thanks{Corresponding author} \email fangxie@uic.edu.cn \\
       \addr Guangdong Provincial Key Laboratory of IRADS\\
             Beijing Normal-Hong Kong Baptist University\\
             Zhuhai 519087, China}

\editor{My editor}
\maketitle

\begin{abstract}
Differential privacy has been proven effective for stochastic gradient descent; however, existing methods often suffer from performance degradation in high-dimensional settings, as the scale of injected noise increases with dimensionality. To tackle this challenge, we propose AdaDPIGU---a new differentially private SGD framework with importance-based gradient updates tailored for deep neural networks. In the pretraining stage, we apply a differentially private Gaussian mechanism to estimate the importance of each parameter while preserving privacy. During the gradient update phase, we prune low-importance coordinates and introduce a coordinate-wise adaptive clipping mechanism, enabling sparse and noise-efficient gradient updates. Theoretically, we prove that AdaDPIGU satisfies $(\varepsilon, \delta)$-differential privacy and retains convergence guarantees. Extensive experiments on standard benchmarks validate the effectiveness of AdaDPIGU. All results are reported under a fixed retention ratio of 60\%. On MNIST, our method achieves a test accuracy of 99.12\% under a privacy budget of $\epsilon = 8$, nearly matching the non-private model. Remarkably, on CIFAR-10, it attains 73.21\% accuracy at $\epsilon = 4$, outperforming the non-private baseline of 71.12\%, demonstrating that adaptive sparsification can enhance both privacy and utility.
\end{abstract}

\begin{keywords}
  Differential privacy, Deep neural networks, Gradient sparsification, Importance update, Adaptive clipping
\end{keywords}

\section{Introduction}
\label{}
Deep learning has achieved remarkable success in tasks such as image recognition~\cite{Lundervold2018AnOO, Zhou2020ARO}, text analysis~\cite{gmail2019}, and recommendation systems~\cite{Lu2012RecommenderS}, largely due to centralized training on large-scale datasets. However, this paradigm raises significant privacy concerns, as user data must be transmitted to a central server for model updates.

Federated Learning~\cite{mcmahan2017communication} addresses this issue by keeping data on local devices and only exchanging model parameters. Nonetheless, studies have shown that model updates alone can leak sensitive information~\cite{fredrikson2015model, Melis2019, NASR2019, phong2017privacy, Song2017MachineLM,  wang2018infer, Zhao2020iDLGID}, indicating that data localization alone is insufficient for rigorous privacy protection.

\begin{figure}[ht]
    \centering
    \includegraphics[width=\textwidth]{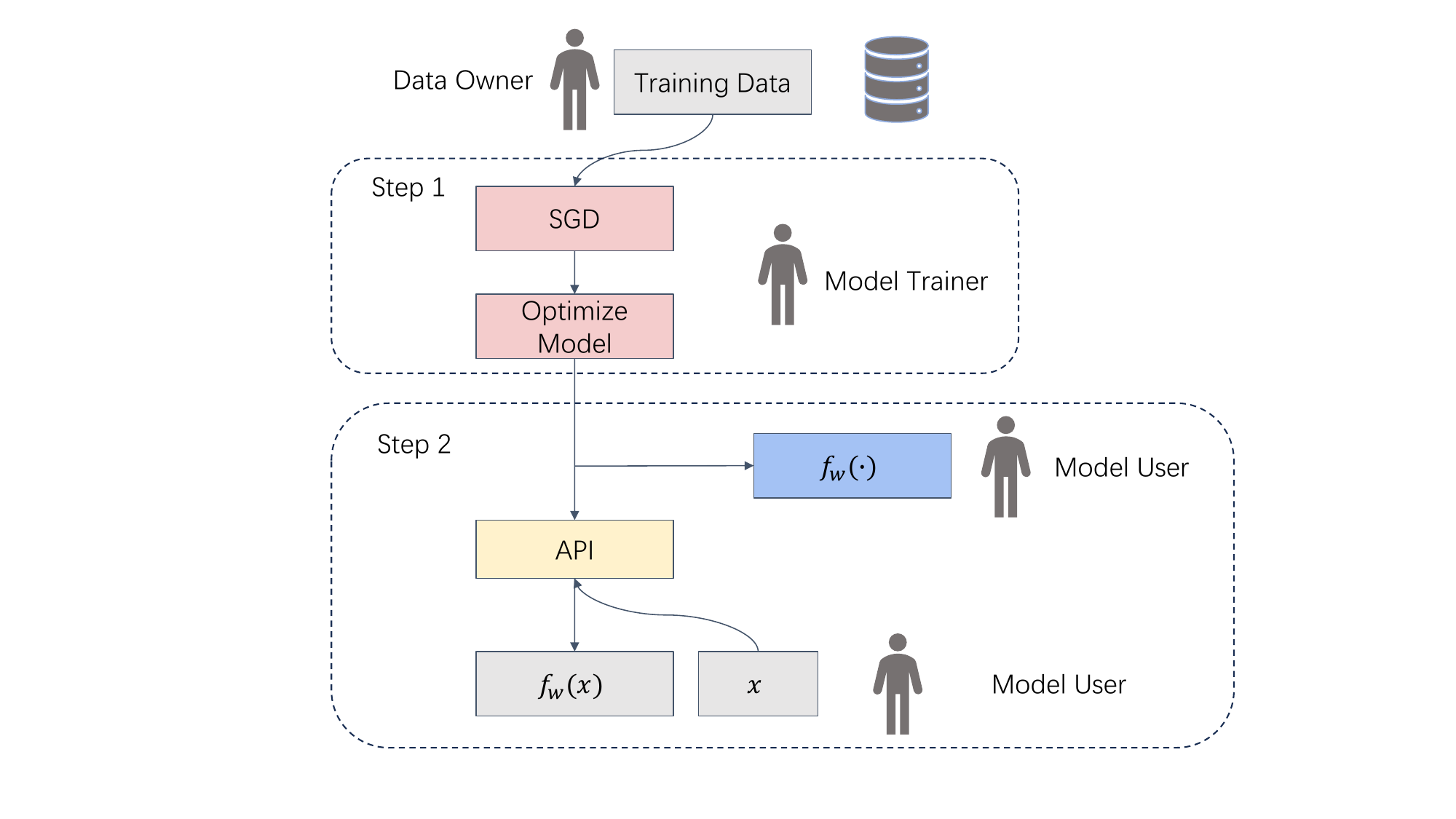} 
    \caption{Deep Learning Pipeline Diagram}
    \label{fig:API}
\end{figure}

As illustrated in Figure~\ref{fig:API}, deep learning systems are vulnerable to privacy threats in both training and deployment. During training, poisoning attacks can degrade performance and leak data~\cite{barreno2006can}, prompting defenses such as data sanitization~\cite{rubinstein2009antidote} and robust learning strategies~\cite{biggio2011bagging, biggio2009multiple}. During deployment, black-box attacks (e.g., model extraction, membership inference~\cite{salem2018ml,  shokri2017membership}) and white-box attacks (e.g., gradient inversion~\cite{Zhao2020iDLGID, zhu2019deep}) threaten user privacy, with recent works extending such threats to federated settings~\cite{geng2021towards, li2021deep, yin2021see, Zhu2020RGAPRG}.

To address privacy concerns in deep learning, existing approaches mainly follow two directions: cryptographic techniques and random perturbation mechanisms. Cryptographic methods include Homomorphic Encryption (HE)~\cite{gentry2009fully, gilad2016cryptonets} and Secure Multi-Party Computation (MPC)~\cite{du2001secure, hastings2019sok, lindell2005secure}. While HE allows computation on encrypted data without decryption and MPC enables secure protocols among untrusted parties, both approaches suffer from high computational overhead, limiting their scalability in deep learning.

Compared to these methods, Differential Privacy (DP)~\cite{dwork2006differential} offers formal guarantees by injecting calibrated noise to obscure the influence of any individual data point. With significantly lower computational costs, DP is better suited for large-scale deep learning applications.

This thesis focuses on privacy protection—distinct from system security—by preventing the leakage of individual data during training through DP mechanisms.

Originally developed for database queries, DP has been successfully adapted to deep learning~\cite{shokri2015privacy}, with Differentially Private Stochastic Gradient Descent (DPSGD)~\cite{abadi2016deep} emerging as a leading method. DPSGD applies per-sample gradient clipping and Gaussian noise addition to ensure formal $(\varepsilon, \delta)$-DP guarantees. However, the injected noise can significantly impair model utility, especially under high-dimensional or long-horizon training regimes.

To understand different noise injection strategies, DP mechanisms are typically classified into three categories: (1) \emph{output perturbation}, which adds noise to the final model~\cite{wu2017bolt} but lacks end-to-end privacy; (2) \emph{objective perturbation}, suitable for convex problems~\cite{chaudhuri2011differentially, iyengar2019towards}; and (3) \emph{gradient perturbation}, used in DPSGD, which adds noise during training updates and is the most widely adopted for deep models.

Despite its popularity, DPSGD faces challenges. First, it injects uniform noise across all gradient dimensions, which scales with model size and impair convergence~\cite{bassily2014private}. Second, the linear accumulation of privacy budget during prolonged training further degrades performance. Additionally, noise is often added to parameters of low importance, thereby weakening the signal for critical gradients~\cite{chen2023differentially}.

To investigate the extent of redundancy in gradient dimensions, we performed a statistical analysis on the MNIST dataset. We found that over 99\% of gradient values lie within $[-0.01, 0.01]$, indicating strong gradient sparsity. These findings, combined with prior studies on gradient pruning~\cite{chen2020understanding}, suggesting that uniform noise injection is overly redundant. These findings motivate us to design a mechanism that dynamically selects important coordinates and focuses the privacy budget on critical parameters, aiming to enhance both utility and privacy.

\begin{figure}[ht]
    \centering
    \includegraphics[width=0.6\textwidth]{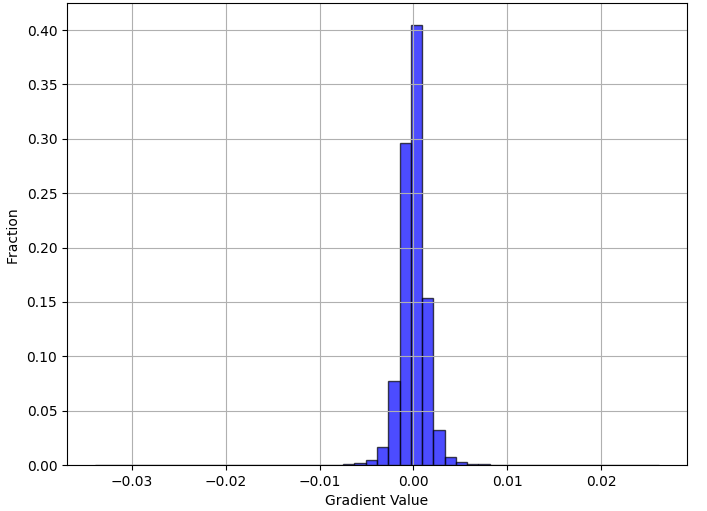}  
    \caption{Gradient Distribution at Epoch 5 (MNIST)}  
    \label{fig:mnist_epoch5}  
\end{figure}
\begin{figure}[htbp]
    \centering
    \begin{subfigure}[b]{0.49\textwidth}
        \includegraphics[width=\textwidth]{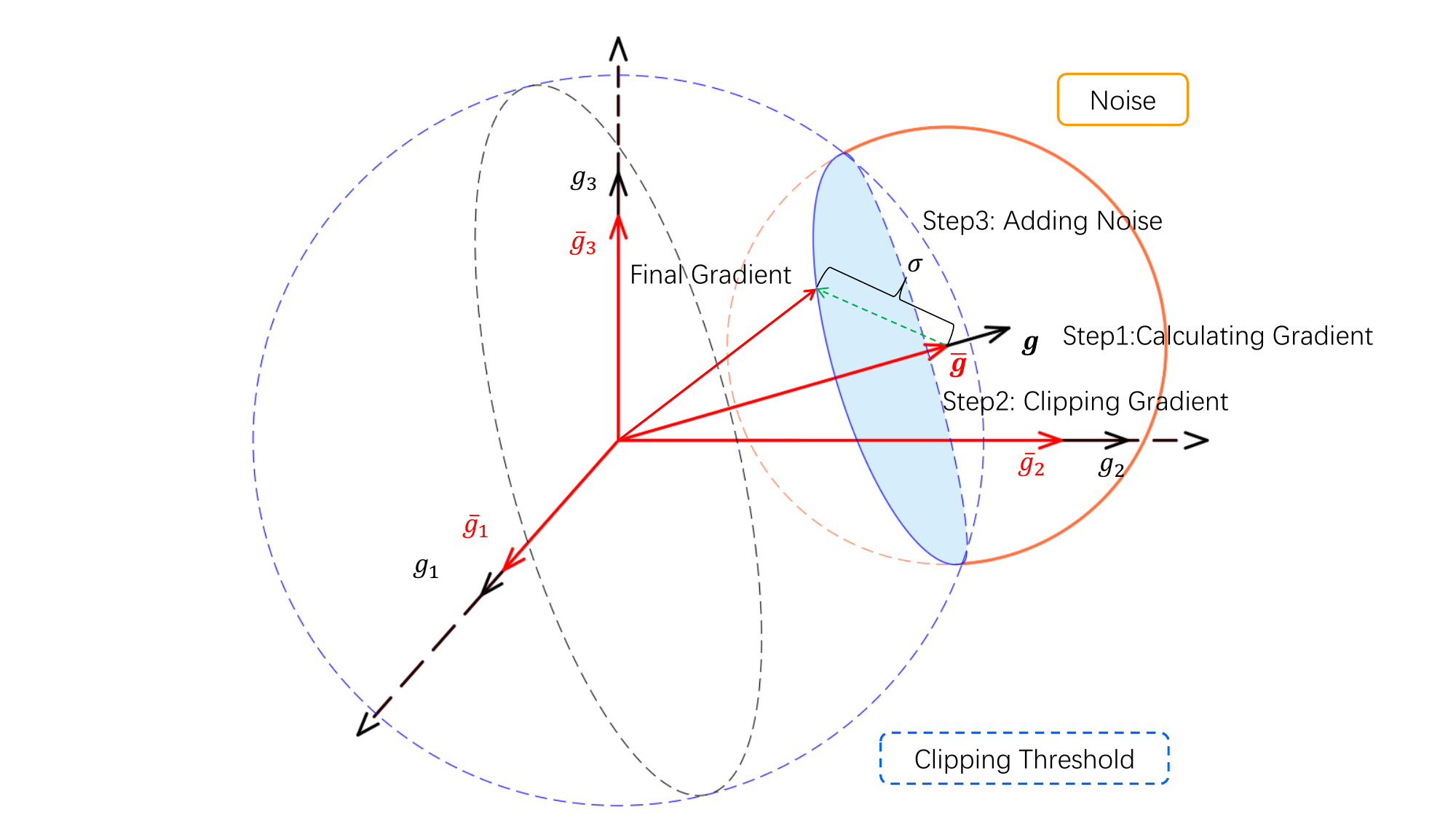}
        \caption{Standard DPSGD gradient update}
        \label{fig:sub1}
    \end{subfigure}
    \hfill
    \begin{subfigure}[b]{0.49\textwidth}
        \includegraphics[width=\textwidth]{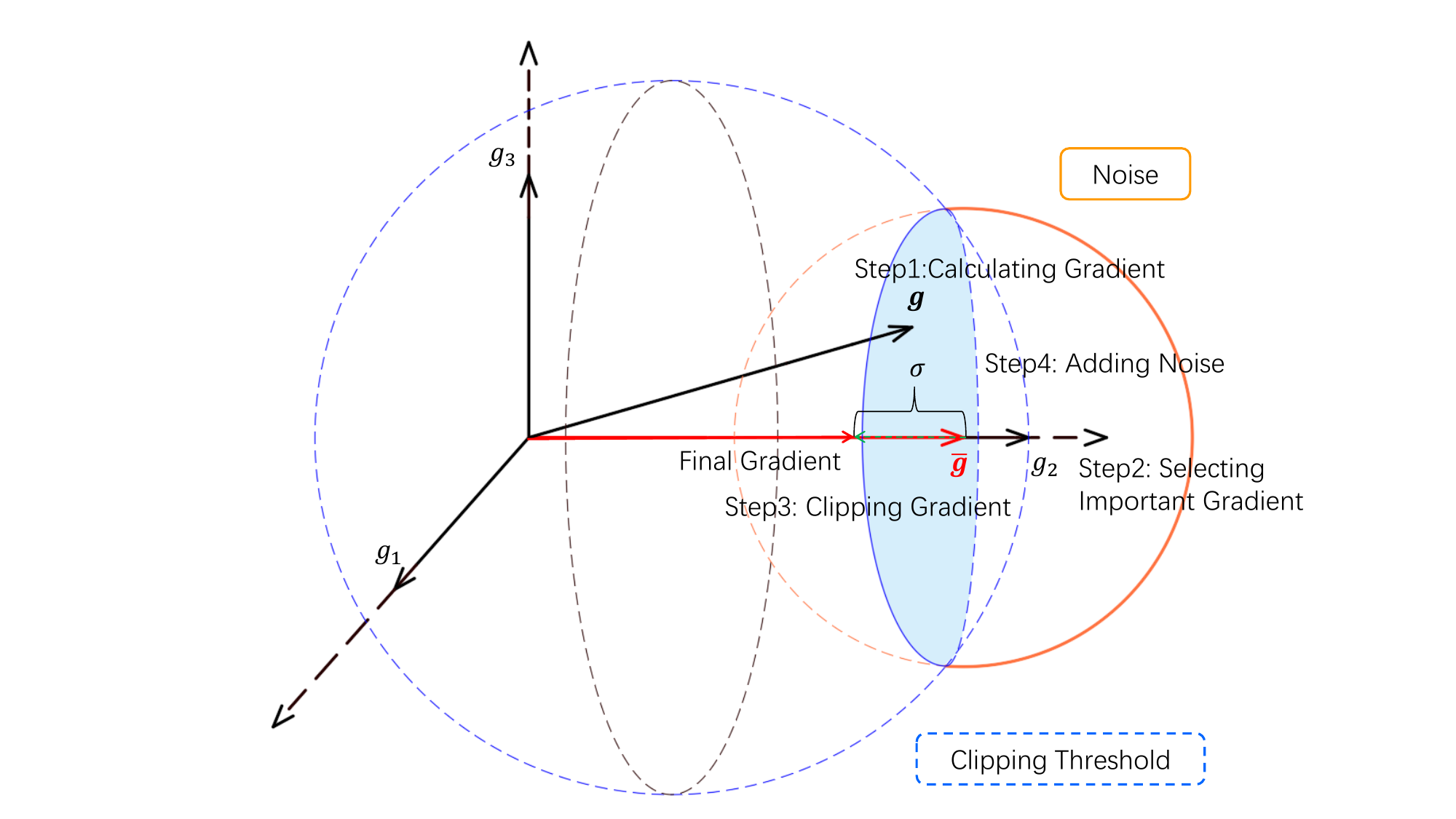}
        \caption{Importance-based sparse gradient update}
        \label{fig:sub2}
    \end{subfigure}
    \caption{Standard DPSGD vs. our AdaDPIGU method for differentially private gradient updates}
    \label{fig:gra3}
\end{figure}

To better appreciate the role of noise injection in DPSGD, we revisit the gradient perturbation pipeline, which consists of the following three steps (see Figure~\ref{fig:sub1}). In the first step (Calculating Gradient), the loss function is differentiated with respect to model parameters during standard SGD to determine update directions. The second step (Clipping Gradient) applies a threshold to rescale gradients that exceed a predefined norm, thereby reducing the impact of individual data samples and mitigating potential privacy leakage. In the final step (Adding Noise), Gaussian noise is added to the clipped gradients, hindering an adversary’s ability to recover the original training data, even if updated gradients are exposed.

The core contributions of this work are as follows:
\begin{itemize}
    \item We propose AdaDPIGU, a new differentially private optimization framework that combines importance-aware gradient selection with adaptive clipping. Unlike standard DPSGD, which clips and perturbs all gradient dimensions uniformly, AdaDPIGU selectively updates only the most informative coordinates at each iteration. The remaining gradients are zeroed out to avoid injecting noise on redundant directions (see Fig.~\ref{fig:sub2}). This sparsity-aware update mechanism significantly reduces the dimensionality of noise injection and preserves critical optimization signals.

    \item We design an adaptive clipping strategy tailored for sparse gradients. By adjusting the clipping threshold based on coordinate-wise statistics, AdaDPIGU avoids over-shrinking the important components. Compared to fixed-norm clipping in DPSGD, this leads to improved signal-to-noise ratio and enhanced training stability, especially in high-dimensional models.

    \item We provide both theoretical and empirical evidence supporting the effectiveness of AdaDPIGU. Our analysis shows that the proposed method satisfies $(\varepsilon, \delta)$-differential privacy guarantees while maintaining convergence. Experiments demonstrate that AdaDPIGU achieves up to 99.12\% accuracy on MNIST using only 60\% of the gradients under $\varepsilon=8$, matching non-private performance and confirming significant gains in both utility and efficiency.
\end{itemize}

\paragraph{Paper Organization} The rest of the paper is organized as follows. Section~\ref{sec:Related Works} reviews related works. Section~\ref{sec:Preliminaries} introduces the preliminaries. In Section~\ref{sec:Our Approach}, we present the AdaDPIGU method and introduce our algorithm. Section~\ref{sec:Theoretical Analysis} provides the privacy analysis and gradient convergence analysis, and Section~\ref{sec:Experiments} presents the experimental results. All proofs are provided in the Appendix.

\section{Related Works}
\label{sec:Related Works}

Differential Privacy~\cite{dwork2006differential} has become the standard method for ensuring data privacy, by limiting the influence of any single sample on the model output and thus protecting data privacy during the training process. DP has been widely applied to the problem of private empirical risk minimization. In deep learning, due to the special challenges posed by model non-convexity and complexity, Abadi et al.~\cite{abadi2016deep} proposed the differentially private stochastic gradient descent algorithm, which has become one of the most widely used methods for privacy preservation. DPSGD limits sensitivity through gradient clipping and adds isotropic Gaussian noise after clipping to effectively perturb the update direction for privacy protection. In addition, the proposed Moment Accounting technique significantly improves the efficiency of privacy budget utilization, enabling tasks such as language modeling to maintain good performance while ensuring privacy protection.

Gradient clipping is a key step in DPSGD to control the sensitivity of each update. However, this operation introduces bias, which may affect the convergence of the optimization process. To further reduce the amount of noise, Pichapati et al.~\cite{pichapati2019adaclip} proposed the AdaCliP method, which dynamically adjusts gradient sensitivity through coordinate-wise adaptive clipping, thereby reducing noise magnitude and improving model performance while preserving privacy. Chen et al.~\cite{chen2020understanding} systematically analyzed the convergence of DPSGD in non-convex settings from a geometric perspective, characterized the bias introduced by gradient clipping, and proposed a perturbation-based correction technique to improve optimization performance. Existing differentially private training methods typically adopt a fixed clipping bound, but it is difficult to set an appropriate value in advance. Andrew et al.~\cite{andrew2021differentially} proposed an adaptive clipping method that privately estimates the quantile of the update norm distribution online, dynamically adjusting the clipping bound and thus eliminating the need for hyperparameter tuning of the clipping threshold.

In recent years, a growing body of research has focused on compressing the gradient space to improve the training utility of DPSGD, aiming to mitigate the performance degradation caused by noise amplification in differentially private training. Abadi et al.~\cite{abadi2016deep} proposed the differentially private linear probing (DP linear probing) method: a full neural network feature extractor is first pretrained on an auxiliary dataset, and then transferred to private data, where only a linear classifier is retrained. By freezing the feature extraction layers on private data and optimizing only in a low-dimensional subspace, this approach significantly reduces the number of parameters to which noise must be added. Tramèr and Boneh~\cite{tramer2020differentially} adopted ScatterNet~\cite{oyallon2019scattering}, using a handcrafted feature extraction network to directly generate input features without training feature layers on private data. Based on this, they trained only a linear classifier on the low-dimensional output features, further reducing the dimensionality and noise impact of differentially private training. Despite differences in methodology, both works essentially compress the gradient space dimension $d$ during differentially private training by freezing most network parameters, thereby significantly weakening the impact of noise injection on model performance.

The aforementioned methods effectively compress the parameter space requiring protection by freezing most of the feature extraction network at the early stage of training. However, such structure-fixing strategies still rely on auxiliary data or a pretraining process, and they constrain the flexible optimization of the network on private data. To address these limitations, recent research has explored dynamically improving the utility of DPSGD during the full training process through subspace compression and gradient sparsification.

Inspired by the empirical observation that neural networks are often heavily over-parameterized, with the number of model parameters far exceeding the minimal scale required to accomplish the task~\cite{lecun1989optimal, han2015learning}, recent studies have proposed improving the utility of differentially private training by compressing the low-dimensional subspace on which gradient updates rely, thereby alleviating the problem of noise amplification.

Based on the empirical observation that gradient updates are often confined to a low-dimensional subspace, Zhou et al.~\cite{zhou2021bypassing} proposed the Projected DPSGD method, which identifies the principal gradient subspace using a small-scale public dataset and projects the noise onto the low-dimensional space to reduce perturbation. Yu et al.~\cite{yu2021large} proposed a method based on low-rank reparametrization, where each weight matrix is decomposed into two small-dimensional gradient carrier matrices and a residual matrix, and designed a reparametrized gradient perturbation (RGP) mechanism that adds noise only in the low-rank subspace. Yu et al.~\cite{yu2021let} further proposed the Gradient Embedding Perturbation (GEP) method, which projects the private gradient of a single sample onto a non-sensitive anchored subspace to obtain low-dimensional embeddings and residuals, and adds noise separately to them. Sha et al.~\cite{sha2023pcdp} introduced the PCDPSGD method, which projects gradients before clipping to compress redundant gradient norms and preserve critical gradient components, thereby improving the convergence of differentially private training. Liu et al.~\cite{liu2024dpdr} proposed the DPDR method, which decomposes the current gradient based on historical noisy gradients, focuses the privacy budget on protecting incremental information, and updates the model using reconstructed gradients to further enhance model utility. In addition, Nasr et al.~\cite{nasr2020improving} proposed the Gradient Encoding and Denoising method, which maps gradients to a smaller vector space to support different noise distributions for differential privacy protection, and utilizes post-processing to further improve model performance. It is worth noting that Vogels et al.~\cite{vogels2019powersgd} proposed the PowerSGD method, which primarily aims to improve communication efficiency by achieving low-rank gradient approximation via power iteration. Although not specifically designed for differential privacy, its low-dimensional compression idea has inspired subsequent subspace optimization techniques in differentially private training.

In addition to optimization strategies based on subspace compression, a series of recent studies have explored introducing sparse updates or parameter pruning to further reduce the impact of noise. Huang et al.~\cite{Huang2020PrivacypreservingLV} proposed a method for privacy preservation via neural network pruning, theoretically proving that pruning is essentially equivalent to adding differential privacy noise to hidden layer activations. Building on this, Zhu and Blaschko~\cite{zhu2021improving} proposed the Random Freeze method, which gradually and randomly freezes parameters during differentially private SGD to generate sparse gradient updates, thereby mitigating the performance degradation caused by noise. Adamczewski and Park~\cite{adamczewski2023differential} further combined neural network pruning with differentially private training, proposing methods that either freeze a subset of parameters before training (parameter freezing) or dynamically select a subset of parameters for updating during training (parameter selection) to reduce the parameter space dimension and enhance the scalability of DPSGD. Both approaches rely on public data for parameter selection, avoiding additional privacy costs. Additionally, Liu et al.~\cite{liu2020fedsel} introduced the FedSel method under a local differential privacy (LDP) setting, where the top-$k$ most contributive dimensions are privately selected in each SGD round and noise is added, alleviating the problem of noise magnitude growing with dimensionality. A delayed gradient accumulation mechanism is also introduced to mitigate the convergence degradation.

\section{Preliminaries}
\label{sec:Preliminaries}
Differential Privacy is a well-established mathematical paradigm that provides formal guarantees to protect individual data confidentiality. The core idea behind this framework is that for individual privacy. The core idea behind this framework is that for any dataset, the output distribution should remain indistinguishable when some data in the dataset is modified or omitted, ensuring that attackers cannot infer private information from queries made on the dataset.

The fundamental objective of differential privacy is to restrict the influence of any single data entry on a query’s output. Put differently, a mechanism is considered differentially private if its responses to two datasets differing by only one record are statistically indistinguishable. To make this concept precise, we define the notion of neighboring datasets as follows

\begin{definition}[Neighboring Datasets~\cite{dwork2006differential}]
Two datasets \( D, D' \in \mathcal{D} \) are considered neighbors if they differ in exactly one element, i.e.,
\begin{equation}
\left| (D \setminus D') \cup (D' \setminus D) \right| = 1.
\end{equation}
\end{definition}

\begin{definition}[$(\varepsilon, \delta)$-Differential Privacy~\cite{dwork2006differential}]
Let \( \mathcal{M} : \mathcal{D} \rightarrow \mathcal{R} \) be a randomized algorithm defined over domain \( \mathcal{D} \) and mapping to range \( \mathcal{R} \).  
For any neighboring inputs \( D, D' \in \mathcal{D} \), and for every measurable event \( S \subseteq \mathcal{R} \),  
the following inequality must be satisfied
\begin{equation}
\mathbb{P}[\mathcal{M}(D) \in S] \leq e^{\varepsilon} \cdot \mathbb{P}[\mathcal{M}(D') \in S] + \delta.
\end{equation}
Then \( \mathcal{M} \) is said to provide \((\varepsilon, \delta)\)-differential privacy.
\end{definition}

Here, \( \varepsilon > 0 \) controls the strength of the privacy guarantee in the worst-case scenario.  
A smaller \( \varepsilon \) indicates stronger privacy protection.  
The parameter \( \delta > 0 \) represents the probability of privacy leakage failure.  
In practice, \( \delta \) is typically set to be negligible~\cite{abadi2016deep, dwork2014algorithmic}.

\begin{lemma}[Post-processing Invariance~\cite{dwork2014algorithmic}]
Let \( \mathcal{M}: \mathcal{D} \rightarrow \mathcal{R} \) be a randomized mechanism that satisfies \((\varepsilon, \delta)\)-differential privacy,  
and let \( f: \mathcal{R} \rightarrow \mathcal{R}' \) be any randomized mapping.  
Then, the composite mechanism \( f \circ \mathcal{M}: \mathcal{D} \rightarrow \mathcal{R}' \) also satisfies \((\varepsilon, \delta)\)-differential privacy.
\end{lemma}

This lemma implies that any post-processing applied to the output of a differentially private mechanism  
(which does not depend on the original data) will not weaken the privacy guarantee.  
This property is especially crucial in practical applications,  
as it allows us to perform multiple rounds of training or analysis  
without incurring additional privacy loss during post-processing.

In the commonly used relaxed definition of differential privacy,  
the Gaussian mechanism is one of the most widely adopted implementations.  
It ensures that the designed algorithm satisfies \((\varepsilon, \delta)\)-differential privacy  
by adding Gaussian-distributed random noise into the function’s result. The precise formulation is stated below.
\begin{definition}[Gaussian Mechanism~\cite{dwork2014algorithmic}]
\label{def:gaussian_mechanism}
Let \( D \in \mathcal{D} \) be a dataset and let \( f: \mathcal{D} \rightarrow \mathbb{R}^d \) be a query function.  
The Gaussian mechanism perturbs the output of \( f \) by adding Gaussian noise calibrated to its sensitivity
\begin{equation}
\mathcal{M}_f(D) = f(D) + \mathcal{N}(\boldsymbol{0}, S_2^2(f)\sigma^2 \boldsymbol{I}_d),
\end{equation}
where \( \sigma \) is the noise multiplier, \( \boldsymbol{I}_d \) is the identity matrix of dimension \( d \), and
$S_2(f) = \sup_{\text{neighboring } D, D'} \|f(D) - f(D')\|_2$
denotes the global \( \ell_2 \)-sensitivity of the mapping \( f \).
\end{definition}
For a mechanism \(\mathcal{M}_f\) that satisfies \((\varepsilon, \delta)\)-DP, we have the relationship
\[
\sigma = S_2(f) \sqrt{\frac{2 \ln{\frac{1.25}{\delta}} }{\varepsilon}}
\]
The added noise is proportional to the $\ell_2$-sensitivity of the function \(f\) and is related to \(\varepsilon\) and \(\delta\). Larger sensitivity values or decreased privacy guarantees lead to more noise. Smaller \(\delta\) also increases the noise. The \((\varepsilon, \delta)\)-differential privacy ensures the privacy bounds of the mechanism.

In deep learning, DP is implemented through gradient perturbation with noise added, commonly achieved using DPSGD. The objective is to prevent adversaries from inferring information about individual samples by observing updates to the model. The process can be broken down into three steps: 

First, for each sample, a gradient perturbation is applied based on its sensitivity. To limit the influence of individual samples, DPSGD applies $\ell_2$-norm clipping to each sample's gradient. The clipped gradient is defined as
\[
\bar{\boldsymbol{g}}_t(\boldsymbol{x}_i) = \frac{\boldsymbol{g}_t(\boldsymbol{x}_i)}{\max\left(1, \frac{\|\boldsymbol{g}_t(\boldsymbol{x}_i)\|_2}{C}\right)},
\]
in which $\|\boldsymbol{g}_t(\boldsymbol{x}_i)\|_2$ measures the $\ell_2$-norm of the gradient for sample $\boldsymbol{x}_i$, and $C > 0$ specifies the norm clipping threshold.  
This operation guarantees that the $\ell_2$-sensitivity of the averaged gradient, when computed over any pair of neighboring datasets $D$ and $D'$, differing by at most one record, does not exceed $C$.

\begin{definition}[$\ell_2$-Sensitivity of Gradient]
Consider two neighboring datasets \( D, D' \in \mathcal{D} \) that differ by a single sample.  
The $\ell_2$-sensitivity of the gradient sum function at iteration \( t \) is defined as
\[
S_2(f) := \max_{D,D' \sim \mathcal{D}} \left\| \sum_{\boldsymbol{x}_i \in D} \boldsymbol{g}_t(\boldsymbol{x}_i) - \sum_{\boldsymbol{x}_i' \in D'} \boldsymbol{g}_t(\boldsymbol{x}_i') \right\|_2.
\]
This sensitivity quantifies the maximum change in the aggregated per-sample gradients when one sample in the dataset is changed.
\end{definition}

Second, after clipping, noise is added to the aggregated gradients in the form of Gaussian noise, which helps obscure the individual contributions of training samples. The privacy guarantee for this mechanism is computed as
\[
\tilde{\boldsymbol{g}}_t \leftarrow \frac{1}{B} \left( \sum_{i \in \mathcal{B}_t} \bar{\boldsymbol{g}}_t + \mathcal{N}(\boldsymbol{0}, \, C^2 \sigma^2 \boldsymbol{I}_d) \right)
\]
Here, \(\sigma\) is the noise multiplier, whose value is calibrated based on the desired privacy parameters \((\varepsilon, \delta)\). This procedure constitutes the well-known Gaussian mechanism in differential privacy.

Since we cannot assume prior knowledge of which individual sample may differ between neighboring datasets, we use a clipping strategy to ensure that the sensitivity of each sample’s gradient contribution is bounded by \(C\). This guarantees that the overall sensitivity of the mechanism remains controlled under the worst-case change of one data point.

Third, at each training step \( t \), the model parameters are updated using the differentially private gradient. The parameter update rule is given by
\[
\boldsymbol{\theta}_{t+1} = \boldsymbol{\theta}_t - \eta \tilde{\boldsymbol{g}}_t,
\]
where \( \tilde{\boldsymbol{g}}_t \) is the noise-perturbed gradient and \( \eta \) is the learning rate.

The model performs a total of \( T \) training steps, where each step involves gradient computation, clipping, noise addition, and parameter update. The privacy cost accumulates over steps, and the training process ensures that the overall privacy loss does not exceed the target privacy budget.

\begin{definition}[$G$-Lipschitz Smoothness]\label{def:glipschitz}
A differentiable function $f: \mathbb{R}^d \rightarrow \mathbb{R}$ is said to be $G$-Lipschitz smooth if its gradient is $G$-Lipschitz continuous, i.e., there exists a constant $G > 0$ such that for all $\boldsymbol{u}, \boldsymbol{v} \in \mathbb{R}^d$,
\[
\| \nabla f(\boldsymbol{u}) - \nabla f(\boldsymbol{v}) \|_2 \leq G \| \boldsymbol{u} - \boldsymbol{v} \|_2.
\]
Equivalently, this condition implies the following upper bound
\[
f(\boldsymbol{u}) \leq f(\boldsymbol{v}) + \langle \nabla f(\boldsymbol{v}), \boldsymbol{u} - \boldsymbol{v} \rangle + \frac{G}{2} \| \boldsymbol{u} - \boldsymbol{v} \|_2^2,
\]
for all $\boldsymbol{u}, \boldsymbol{v} \in \mathbb{R}^d$.
\end{definition}

\section{Our Approach}\label{sec:Our Approach}

Stochastic Gradient Descent remains the foundation of modern deep learning optimization. To ensure privacy, Abadi et al.~\cite{abadi2016deep} proposed DPSGD, which enforces $(\varepsilon, \delta)$-differential privacy by clipping per-sample gradients and injecting Gaussian noise. However, DPSGD often suffers from slower convergence and reduced utility, primarily due to two factors: (1) uniform noise addition across all parameter dimensions, which scales with model size and weakens critical learning signals~\cite{bassily2014private}; and (2) fixed clipping thresholds that fail to adapt to gradient sparsity or distributional changes.

While prior works have explored adaptive clipping and sparsity-aware mechanisms, they often rely on public data, introduce instability via random masking, or delay updates due to rigid structural constraints. To address these challenges, we propose AdaDPIGU, a differentially private training framework that combines importance-aware gradient selection with adaptive noise control.

By focusing updates on the most informative coordinates, AdaDPIGU reduces unnecessary perturbations and improves the signal-to-noise ratio. Importantly, the selected gradients often exhibit smaller magnitudes, which allow for tighter and more stable clipping bounds. Implementing such selective updates under differential privacy requires careful coordination of importance estimation, dynamic clipping, and noise calibration.

To this end, AdaDPIGU integrates the following components:
\begin{itemize}
    \item \textbf{Importance-Based Masking with Progressive Unfreezing:} A binary mask is applied early in training to update only the most critical parameters. As training progresses, the mask is gradually relaxed to restore full model capacity, reducing early-stage noise where gradients are unstable.
    
    \item \textbf{Adaptive Clipping:} Clipping thresholds are dynamically set based on the distribution of selected gradients. This ensures effective noise scaling that matches the sparsity structure and gradient magnitudes.

    \item \textbf{Progressive Sparsification:} The retention ratio is gradually increased over training, allowing privacy noise to concentrate on fewer parameters at early stages and thereby improving privacy-utility trade-off over time.
\end{itemize}

We flatten all trainable parameters in the deep neural network into a $d$-dimensional vector. 
Let $\boldsymbol{\theta} = [\theta_1, \theta_2, \dots, \theta_d]^\top \in \mathbb{R}^d$ denote the flattened parameter vector, where each scalar component $\theta_j$ corresponds to a specific trainable weight or bias in the model.

Given a loss function $\mathcal{L}(\boldsymbol{\theta})$, the gradient vector $\boldsymbol{g}$ is defined as
\[
\boldsymbol{g} = \nabla \mathcal{L}(\boldsymbol{\theta}) = 
\left[
\frac{\partial \mathcal{L}}{\partial \theta_1},\ 
\frac{\partial \mathcal{L}}{\partial \theta_2},\ 
\dots,\ 
\frac{\partial \mathcal{L}}{\partial \theta_d}
\right]^\top,
\]
each component represents the partial derivative of the loss with respect to the corresponding parameter, indicating how sensitive the loss is to changes in that parameter. This vector is used to guide parameter updates during training.

\begin{figure}[htbp]
    \centering
    \includegraphics[width=\textwidth]{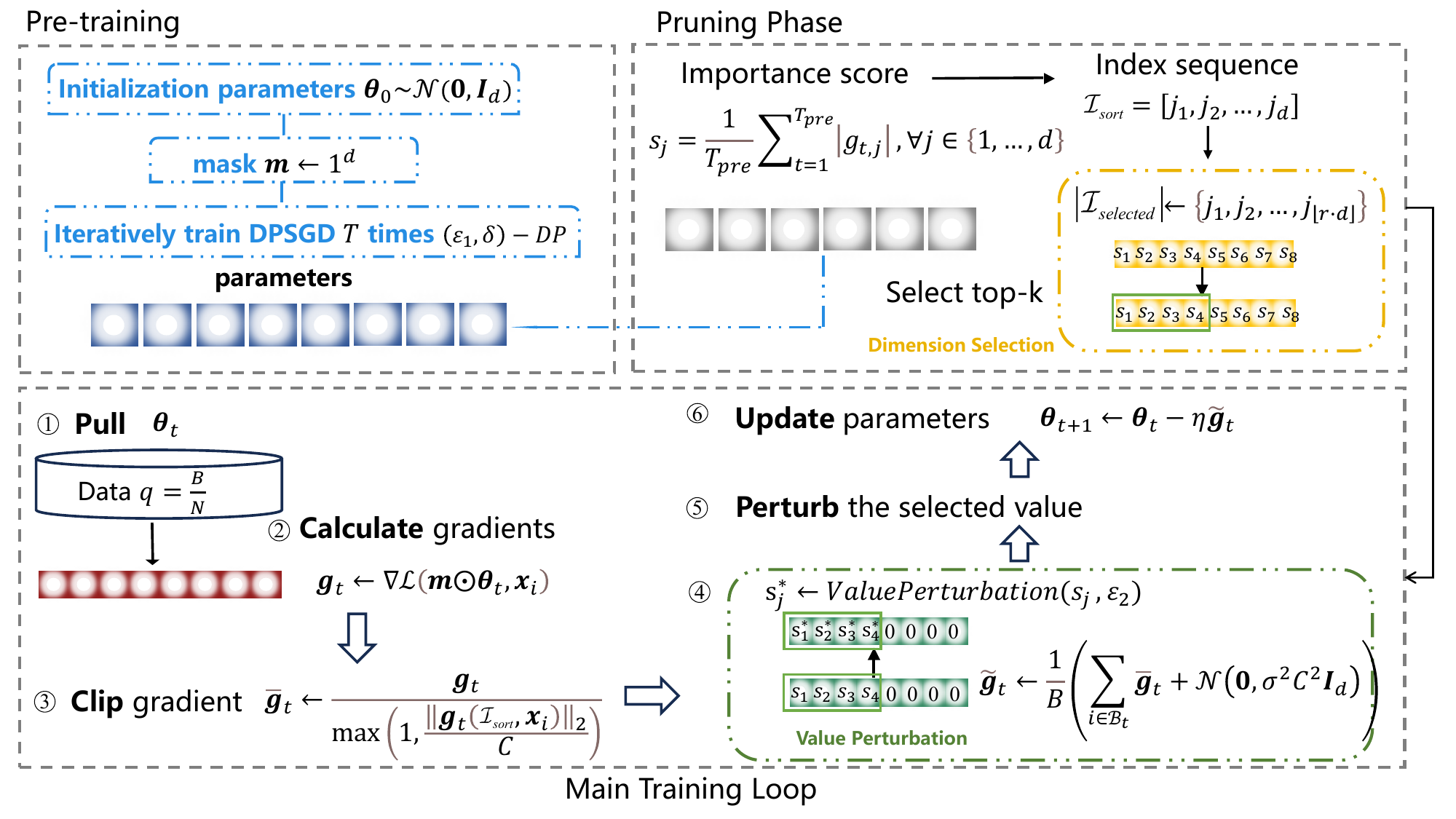}
    \caption{Overview of the proposed AdaDPIGU training pipeline.}
    \label{fig:dpigu_pipeline}
\end{figure}
Figure~\ref{fig:dpigu_pipeline} illustrates the overall pipeline of AdaDPIGU. The training process starts with a pre-training phase, where the model parameters are trained under differential privacy to capture the distribution of important weights. This is followed by a pruning phase, where an importance score is computed for each parameter based on the pre-training trajectory, and only the top-\(k\) most important parameters are retained for subsequent updates. Finally, in the Main Training Loop, only the selected important gradients are computed, clipped, and perturbed with noise, significantly reducing noise addition compared to standard DPSGD. This design effectively improves training efficiency and enhances the privacy-utility trade-off.

In the first stage of the AdaDPIGU training framework (pre-training and gradient importance evaluation), we compute the cumulative gradient magnitudes for each parameter dimension to measure their contribution to optimization across multiple iterations, thereby constructing a parameter importance score. Based on this score, AdaDPIGU constructs a sparse gradient mask to retain critical gradients and mask redundant gradients, effectively compressing the update space and improving noise utilization.

We begin by performing $T_{\mathrm{pre}}$ iterations of standard DPSGD on the original model structure to optimize the parameters and simultaneously accumulate parameter-level importance statistics. Specifically, we compute the importance score $s_j$ for each parameter dimension $j$ as follows
\[
s_j = \frac{1}{T_{\mathrm{pre}}} \sum_{t=1}^{T_{\mathrm{pre}}} |g_{t,j}|, \quad \forall j \in \{1, \dots, d\},
\]
where $g_{t,j}$ denotes the $j$-th coordinate of the average (clipped) gradient at iteration $t$. This scalar score $s_j$ reflects the average gradient magnitude and serves as a proxy for parameter importance. We collect all scores into an importance vector $\boldsymbol{s} = (s_1, s_2, \dots, s_d)^\top \in \mathbb{R}^d$, which is used for top-$k$ pruning in subsequent training stages.

We then sort all coordinates in descending order of their importance scores to obtain the ranked index sequence
\[
\mathcal{I}_{\mathrm{sort}} = [j_1, j_2, \dots, j_d] \quad \text{such that} \quad s_{j_1} \geq s_{j_2} \geq \cdots \geq s_{j_d}.
\]
Let \( r \in (0, 1] \) be the parameter selection ratio, and define the selected index set as
\[
\mathcal{I}_s := \left\{ j_1, j_2, \dots, j_{\lfloor r \cdot d \rfloor} \right\},
\]
which includes the top \(r\) fraction of coordinates with the highest importance scores. These selected coordinates are retained and updated during subsequent training, while the remaining coordinates are masked out to reduce noise injection.
  
The remaining coordinates that are excluded from updates form the non-selected index set
\[
\mathcal{I}_{\text{ns}} := \left\{ j_{\lfloor r \cdot d \rfloor + 1}, j_{\lfloor r \cdot d \rfloor + 2}, \dots, j_d \right\}.
\]
Based on $\mathcal{I}_s$, we construct a binary importance-based mask vector $\boldsymbol{m} = (m_1, m_2, \dots, m_d) \in \{0,1\}^d$, where
\[
m_j =
\begin{cases}
1, & \text{if } j \in \mathcal{I}_s, \\
0, & \text{otherwise}.
\end{cases}
\]
Here, $\mathcal{I}_s$ denotes the set of important coordinates that will be retained and updated, while the remaining coordinates $\mathcal{I}_{\text{ns}}$ are masked out and excluded from gradient updates in the subsequent training process.

To avoid overly aggressive pruning that may harm model capacity, we adopt a progressive unfreezing strategy that gradually releases masked coordinates during training. Specifically, instead of using a fixed mask $\boldsymbol{m}$ throughout the entire training process, we define a release schedule $\{r_t\}_{t=1}^T$, where each \(r_t \in [r, 1)\) denotes the \emph{retention ratio} at training step \(t\), controlling the fraction of coordinates allowed to participate in updates.

At the beginning of training, only the top \(r \cdot d\) coordinates (i.e., those with the highest importance scores) are active. As training progresses, more coordinates from the non-selected set $\mathcal{I}_{\text{ns}}$ are incrementally reintroduced into the update process according to the schedule $\{r_t\}$. This yields a dynamic binary mask $\boldsymbol{m}^{(t)} \in \{0,1\}^d$ that evolves over time, expanding the set of trainable parameters.

The unfreezing process is managed by a mask scheduling mechanism, which activates the first \(\lfloor r_t \cdot d \rfloor\) coordinates in the importance ranking $\mathcal{I}_{\mathrm{sort}}$ at each step \(t\). This approach enables the model to initially focus on the most critical parameters, while gradually restoring full model capacity, thus balancing privacy-preserving sparsity and expressive model updates.

To determine the parameter importance ranking, we perform a DPSGD-based pretraining phase to accumulate per-parameter importance scores. Specifically, the absolute values of each parameter's gradient are recorded and averaged across \(T_{\mathrm{pre}}\) iterations. These scores are used to rank all parameter dimensions in descending order. By retaining only the top-\(r\) fraction of parameters at each step according to this ranking, we construct a time-varying binary mask $\boldsymbol{m}^{(t)}$ that specifies which coordinates are included in the update. The full procedure is summarized in Algorithm~\ref{alg:importance_mask}.

\begin{algorithm}[ht]
\caption{Privacy-Preserving Importance Scoring and Mask Generation}
\label{alg:importance_mask}
\begin{algorithmic}[1]
\State \textbf{Input:} Model parameters $\boldsymbol{\theta}_0 \in \mathbb{R}^d$; pretraining steps $T_{\mathrm{pre}}$; parameter selection rate $r \in (0, 1]$
\State Initialize importance scores: $\boldsymbol{s} \gets \boldsymbol{0}^d$
\For{$t = 1$ to $T_{\mathrm{pre}}$}
    \State Run DPSGD on batch $\mathcal{B}_t$ to obtain clipped gradient $\boldsymbol{g}_t \in \mathbb{R}^d$
    \For{$j = 1$ to $d$}
        \State $s_j \gets s_j + |g_{t,j}|$
    \EndFor
\EndFor

\State Normalize importance scores: $s_j \gets s_j / T_{\mathrm{pre}}$ for all $j$
\State Sort indices: $\mathcal{I}_{\mathrm{sort}} \gets [j_1, \dots, j_d]$ such that $s_{j_1} \geq s_{j_2} \geq \cdots \geq s_{j_d}$
\State Compute threshold: $k \gets \lfloor r \cdot d \rfloor$
\State Define selected set: $\mathcal{I}_s \gets \{j_1, j_2, \dots, j_k\}$
\For{$j = 1$ to $d$}
    \State $m_j \gets \begin{cases}
        1, & \text{if } j \in \mathcal{I}_s \\
        0, & \text{otherwise}
    \end{cases}$
\EndFor
\State \textbf{Output:} Importance vector $\boldsymbol{s} \in \mathbb{R}^d$; pruning mask $\boldsymbol{m} \in \{0, 1\}^d$
\end{algorithmic}
\end{algorithm}

To overcome the limitations of uniform perturbation in deep models, we propose an adaptive gradient perturbation framework that dynamically calibrates sensitivity based on coordinate-wise statistics. Traditional DP mechanisms such as DPSGD rely on global clipping and isotropic noise addition, which fail to account for the heterogeneity in gradient scale and informativeness across dimensions—especially in high-dimensional deep networks.

Our framework incorporates coordinate-wise statistical estimation, per-sample pruning, adaptive clipping, and noise injection with scale-aware recovery, enabling a more fine-grained and informative application of privacy noise. This design not only tightens sensitivity bounds but also reduces unnecessary distortion to important gradient components.

\paragraph{Standardizing and Scaling}
To enable fine-grained sensitivity control, we estimate coordinate-wise gradient statistics over recent training steps. Let \(\boldsymbol{\alpha}, \boldsymbol{\beta} \in \mathbb{R}^d\) denote the moving averages of the mean and variance of the gradients, respectively. For each sample \(\boldsymbol{x}_i\), the corresponding gradient vector \(\boldsymbol{g}(\boldsymbol{x}_i) \in \mathbb{R}^d\) is standardized as
\[
\tilde{\boldsymbol{g}}(\boldsymbol{x}_i) = \frac{ \boldsymbol{g}(\boldsymbol{x}_i) - \boldsymbol{\alpha} }{ \sqrt{\boldsymbol{\beta}} + \mu \cdot \boldsymbol{1}},
\]
where $\mu > 0$ denotes a small constant introduced to enhance numerical stability.

Unlike classical normalization that reduces global variance, our method adaptively rescales each coordinate according to its dynamic range. This coordinate-wise adaptive clipping mechanism enables tighter sensitivity control under \((\varepsilon, \delta)\)-DP while preserving important gradient signals—especially in high-dimensional settings with heterogeneous gradient scales.

\paragraph{Pruning via top-$k$ Selection}
After standardization, we perform coordinate-wise pruning on each sample's gradient. Specifically, for each sample $\boldsymbol{x}_i$, we construct a binary mask $\boldsymbol{m} \in \{0,1\}^d$ as follows
\[
m_j =
\begin{cases}
1, & \text{if } \left| \tilde{\boldsymbol{g}}_j(\boldsymbol{x}_i) \right| \geq \tau \\
0, & \text{otherwise}
\end{cases}
\quad \text{for all } j \in \{1, \dots, d\}.
\]
Here, \(\tau\) is a sample-specific threshold such that only the largest \(r\) fraction of coordinates in \( |\tilde{\boldsymbol{g}}(\boldsymbol{x}_i)| \) are retained, where \( r \in (0,1) \) is the \emph{retention ratio}. This pruning strategy enables per-sample sparsification, focusing the privacy budget on the most salient gradient components.

\paragraph{Calculating Clipping}
To ensure bounded sensitivity, we apply vector-level \(\ell_2\) norm clipping to each standardized gradient vector. Specifically, the clipped gradient for sample \(\boldsymbol{x}_i\) is computed as
\[
\bar{\boldsymbol{g}}(\boldsymbol{x}_i) \leftarrow \frac{ \tilde{\boldsymbol{g}}(\boldsymbol{x}_i) }{ \max\left(1, \frac{ \| \tilde{\boldsymbol{g}}(\boldsymbol{x}_i) \|_2 }{C} \right) }.
\]
This guarantees that the clipped gradient satisfies $\| \bar{\boldsymbol{g}}(\boldsymbol{x}_i) \|_2 \leq C$. By scaling only when the norm exceeds the threshold, this step ensures that the total sensitivity across all active (unmasked) coordinates remains within a controlled bound.

If coordinate-wise pruning is applied beforehand, the clipping is performed only on the active coordinates, yielding
\[
\bar{\boldsymbol{g}}(\boldsymbol{x}_i) \leftarrow \frac{ \boldsymbol{m} \odot \tilde{\boldsymbol{g}}(\boldsymbol{x}_i) }{ \max\left(1, \frac{ \| \boldsymbol{m} \odot \tilde{\boldsymbol{g}}(\boldsymbol{x}_i) \|_2 }{C} \right) }.
\]
This formulation ensures that sensitivity is bounded within the subspace defined by the pruning mask \(\boldsymbol{m}\), thereby focusing privacy protection on the informative gradient components.

\paragraph{Adding Noise}
To enforce \((\varepsilon, \delta)\)-differential privacy, we inject noise into the aggregated gradient after clipping. Specifically, given a mini-batch \(\mathcal{B}_t\) of size \(B\), we compute the noisy gradient as
\[
\tilde{\boldsymbol{g}}_t \leftarrow \frac{1}{B} \left( \sum_{i \in \mathcal{B}_t} \bar{\boldsymbol{g}}(\boldsymbol{x}_i) + \mathcal{N}(\boldsymbol{0}, C^2 \sigma^2 \boldsymbol{I}_d) \right),
\]
where \(\bar{\boldsymbol{g}}(\boldsymbol{x}_i)\) is the clipped (and optionally pruned) gradient, \(C\) is the clipping norm bound, and \(\sigma\) controls the noise scale.

The noise is sampled independently for each coordinate and scaled according to the worst-case sensitivity \(C\), ensuring that the update satisfies DP guarantees. This design aligns with prior work such as AdaCliP, which adaptively adds noise based on per-coordinate statistics. Compared to uniform perturbation over all coordinates, this approach better preserves informative dimensions and reduces unnecessary distortion in sparse or heterogeneous gradient settings.

\paragraph{Restoring Gradients}
Finally, after applying Gaussian noise in the standardized space, we restore the gradients back to their original scale using the coordinate-wise mean and variance estimators. The final recovered gradient is computed as
\[
\hat{\boldsymbol{g}}_t = \tilde{\boldsymbol{g}}_t \odot \left( \sqrt{\boldsymbol{\beta}} + \mu \cdot \boldsymbol{1}\right) + \boldsymbol{\alpha},
\]
where $\tilde{\boldsymbol{g}}_t$ is the noisy gradient in the standardized space, $\boldsymbol{\alpha}$ and $\boldsymbol{\beta}$ denote the estimated coordinate-wise mean and variance of the original gradients, and $\mu$ is a small constant to ensure numerical stability.

This restoration step reverses the earlier standardization transformation, ensuring that the final update reflects the original gradient scale, with noise appropriately adapted to each coordinate’s statistical sensitivity.

\paragraph{Updating Mean and Variance}
After each training iteration, we update the coordinate-wise estimates of the mean and variance of the gradients. Specifically, the updates are given by
\[
\boldsymbol{\alpha}_{t+1} \leftarrow \gamma_1 \cdot \boldsymbol{\alpha}_t + (1 - \gamma_1) \cdot \hat{\boldsymbol{g}}_t,
\]
\[
\boldsymbol{\beta}_{t+1} \leftarrow \gamma_2 \cdot \boldsymbol{\beta}_t + (1 - \gamma_2) \cdot (\hat{\boldsymbol{g}}_t - \boldsymbol{\alpha}_t)^2,
\]
where \(\gamma_1, \gamma_2 \in (0, 1)\) are momentum hyperparameters that control the temporal smoothing of the statistics.

Unlike traditional DPSGD, which applies global norm clipping and uniform noise injection, our method introduces a coordinate-wise adaptive perturbation mechanism tailored to gradient structure. Operating in the standardized space, AdaDPIGU achieves finer-grained sensitivity control by prioritizing informative gradient dimensions and suppressing irrelevant ones, thereby improving the signal-to-noise ratio under differential privacy. As an enhanced variant of the AdaDPIGU pipeline, AdaDPIGU preserves rigorous $(\varepsilon, \delta)$-DP guarantees while offering improved utility, especially in high-dimensional or sparse training settings.

\begin{algorithm}[]
\caption{AdaDPIGU: Adaptive DPSGD with Importance-Based Updating}
\label{alg:adadpigu}
\begin{algorithmic}[1]
\State \textbf{Input:} Training dataset $D = \{\boldsymbol{x}_i\}_{i=1}^N$; objective function $\mathcal{L}(\boldsymbol{\theta})$; step size $\eta$; noise scale $\sigma$; mini-batch size $B$; norm bound $C$; retention ratio $r$; total iterations $T$; exponential decay rates $\gamma_1, \gamma_2$; initialization values $\boldsymbol{\alpha}_0, \boldsymbol{\beta}_0$; positive constant $\mu > 0$
\State \textbf{Importance Score Accumulation and Pruning Mask Generation}
\State \hspace{1em} Use Algorithm~\ref{alg:importance_mask} to obtain importance scores $\boldsymbol{s}$ and pruning mask $\boldsymbol{m}$

\State \textbf{Adaptive DP Training with Masked Sparse Gradient Updates}
\For{$t = 1$ to $T$}
    \State Sample mini-batch $\mathcal{B}_t \subset \mathcal{D}$ of size $B$
    \For{each $(\boldsymbol{x}_i) \in \mathcal{B}_t$}
        \State Compute masked gradient: $\boldsymbol{g}_t(\boldsymbol{x}_i) = \nabla_{\boldsymbol{\theta}} \ell(\boldsymbol{m} \odot \boldsymbol{\theta}_t, \boldsymbol{x}_i)$
        \State Standardize: $\tilde{\boldsymbol{g}}_t(\boldsymbol{x}_i) = \frac{ \boldsymbol{g}_t(\boldsymbol{x}_i) - \boldsymbol{\alpha}_t }{ \sqrt{\boldsymbol{\beta}_t} + \mu \cdot \boldsymbol{1} }$
        \State Clip: $\bar{\boldsymbol{g}}_t(\boldsymbol{x}_i) = \frac{ \tilde{\boldsymbol{g}}_t(\boldsymbol{x}_i) }{ \max\left(1, \frac{ \|\tilde{\boldsymbol{g}}_t(\boldsymbol{x}_i)\|_2 }{C} \right)}$
    \EndFor
    \State Aggregate with noise
    \[
    \bar{\boldsymbol{g}}_t = \frac{1}{B} \left( \sum_{i \in \mathcal{B}_t} \bar{\boldsymbol{g}}_t(\boldsymbol{x}_i) + \mathcal{N}(\boldsymbol{0}, C^2 \sigma^2 I_d) \right)
    \]
    \State Restore: $\hat{\boldsymbol{g}}_t = \bar{\boldsymbol{g}}_t \odot (\sqrt{\boldsymbol{\beta}_t} + \mu \cdot \boldsymbol{1}) + \boldsymbol{\alpha}_t$
    \State Update parameters: $\boldsymbol{\theta}_{t+1} \leftarrow \boldsymbol{\theta}_t - \eta \cdot \hat{\boldsymbol{g}}_t$
    \State Update statistics
    \[
    \begin{aligned}
    \boldsymbol{\alpha}_{t+1} &\leftarrow \gamma_1 \boldsymbol{\alpha}_t + (1 - \gamma_1)\hat{\boldsymbol{g}}_t \\
    \boldsymbol{\beta}_{t+1} &\leftarrow \gamma_2 \boldsymbol{\beta}_t + (1 - \gamma_2)(\hat{\boldsymbol{g}}_t - \boldsymbol{\alpha}_t)^2
    \end{aligned}
    \]
\EndFor
\State \textbf{Output:} Final parameters $\boldsymbol{\theta}_T$ satisfying $(\varepsilon, \delta)$-DP
\end{algorithmic}
\end{algorithm}

We summarize the proposed AdaDPIGU framework in Algorithm~\ref{alg:adadpigu}, which describes the main training procedure following importance-based pruning.

In each training step, the model first applies a binary pruning mask $\boldsymbol{m} \in \{0,1\}^d$ derived from Algorithm~\ref{alg:importance_mask} to enforce structural sparsity on parameters and gradients. For each sample in a mini-batch, the masked gradient is standardized by subtracting a running estimate of the mean $\boldsymbol{\alpha}_t$ and dividing by the square root of the running variance $\boldsymbol{\beta}_t$, plus a small constant $\mu$ for numerical stability.

After standardization, we retain only the $r$-fraction of coordinates with the largest magnitudes in each gradient vector, where $r \in (0,1)$ is the retention ratio that determines the level of sparsity. The retained values are then clipped in $\ell_2$ norm to ensure that the gradient sensitivity is bounded by $C$. These clipped gradients are averaged across the batch and perturbed by Gaussian noise with variance proportional to $C^2 \sigma^2$, where $\sigma$ is the noise multiplier.

The final noisy gradient is recovered by reversing the standardization transformation and is subsequently applied to perform parameter updates. Moving statistics $\boldsymbol{\alpha}_t$ and $\boldsymbol{\beta}_t$ are also updated using exponential moving averages controlled by momentum factors $\gamma_1$ and $\gamma_2$.

Overall, this adaptive training procedure allows the model to focus the privacy budget on a fixed subset of important coordinates, thereby reducing noise accumulation and improving the utility of differentially private training.

\section{Theoretical Analysis}
\label{sec:Theoretical Analysis}
\subsection{Privacy Analysis of DP Optimizers}

Differential privacy has become a standard mechanism for protecting individual data during deep learning training. In the context of optimization, it is typically enforced by perturbing per-sample gradients with carefully calibrated noise, such that the model's behavior remains insensitive to the presence or absence of any single data point.

More formally, a randomized algorithm $M$ satisfies $(\varepsilon, \delta)$-DP if, for any two neighboring datasets differing in one element, the output distributions of $M$ on these datasets are close in the sense of $(\varepsilon, \delta)$-indistinguishability~\cite{dwork2006differential}. One important property of differential privacy is that it is preserved under post-processing: if $M$ is $(\varepsilon, \delta)$-DP, then any deterministic or randomized function applied to $M$’s output does not weaken its privacy guarantee.

However, standard DP suffers from limited composability, making it difficult to tightly track the cumulative privacy loss over multiple training iterations. To address this, we adopt \emph{R\'enyi Differential Privacy (RDP)}~\cite{mironov2017renyi}, a relaxation of DP based on R\'enyi divergence, which enables more accurate analysis of privacy loss under repeated composition of differentially private mechanisms, especially in stochastic gradient descent settings.

\begin{definition}[RDP {\cite{mironov2017renyi}}]
Let $M: \mathcal{D} \rightarrow \mathcal{R}$ be a randomized mechanism. For any pair of adjacent datasets $d, d' \in \mathcal{D}$ and any subset of outputs $\mathcal{S} \subseteq \mathcal{R}$. Then for all $\alpha \in (1, \infty)$, the following holds
\[
D_\alpha(M(d) \parallel M(d')) = \frac{1}{\alpha - 1} \log \mathbb{E}_{\boldsymbol{\theta} \sim M(d')}
\left[ \left( \frac{\Pr[M(d) = \boldsymbol{\theta}]}{\Pr[M(d') = \boldsymbol{\theta}]} \right)^\alpha \right] \leq \varepsilon.
\]
We say that $M$ satisfies $(\alpha, \varepsilon)$-RDP if $D_\alpha(M(d) \parallel M(d')) \leq \varepsilon$ for all such $d, d'$.
\end{definition}

This definition provides a tunable and more fine-grained measure of privacy loss than standard $(\varepsilon, \delta)$-DP. In particular, RDP enables tight privacy composition and accounting over multiple steps of training. Notably, as $\alpha \to 1$, RDP converges to the standard differential privacy definition.

To enable conversion from RDP to standard $(\varepsilon, \delta)$-DP guarantees in practical applications, we adopt the following proposition

\begin{proposition}[RDP to $(\varepsilon, \delta)$-DP Conversion {\cite{mironov2017renyi}}]\label{From RDP to DP}
If a mechanism $M$ satisfies $(\alpha, \varepsilon)$-RDP for some $\alpha > 1$, then for any $\delta \in (0, 1)$, it also satisfies
\[
\left( \varepsilon + \frac{\log(1/\delta)}{\alpha - 1}, \delta \right)\text{-DP}.
\]
\end{proposition}

This conversion allows us to leverage the tighter composition properties of RDP during training, while still reporting final privacy guarantees in the standard $(\varepsilon, \delta)$ format for compatibility with DP benchmarks and regulatory requirements.

DPSGD ensures privacy by clipping each individual gradient to bound sensitivity and adding random noise to obscure specific contributions. This mechanism limits the influence of any single data point on the final model, thereby preventing privacy leakage through gradient updates.

In practice, DPSGD achieves $(\varepsilon, \delta)$-differential privacy by performing per-sample gradient clipping and Gaussian perturbation at each step. The privacy loss accumulates over multiple training iterations and can be tightly bounded under the Rényi Differential Privacy framework, as formalized below:

\begin{theorem}[Privacy Guarantee of DPSGD~\cite{abadi2016deep}]\label{the:DPSGD}
Consider a dataset of size $N$. In each iteration, a mini-batch of size $B$ is sampled uniformly at random with sampling rate $q = B/N$. Each per-sample gradient is clipped to norm bound $C$ and perturbed by Gaussian noise. After $T$ training steps, the overall mechanism satisfies $(\varepsilon, \delta)$-DP if the noise multiplier $\sigma$ satisfies
\[
\sigma \geq \frac{2q \sqrt{T \ln(1/\delta)}}{\varepsilon}.
\]
\end{theorem}
The proof is given in~\ref{proof:DPSGD}.
This bound is derived under the Rényi Differential Privacy composition framework and quantifies the required noise level to ensure end-to-end differential privacy after $T$ steps.

To further reduce the influence of individual samples on the model update, a subsampling mechanism is employed. This mechanism not only reduces the probability that a sample is observed in each iteration but also weakens its impact on the overall privacy loss.

The following theorem characterizes the privacy amplification effect of subsampling,
\begin{theorem}[Privacy Amplification by Subsampling~\cite{wang2019subsampled}] 
If a mechanism $M$ satisfies $(\varepsilon, \delta)$-DP on the full dataset, then the subsampled mechanism $M' = M \circ \texttt{Subsample}$ satisfies $(\varepsilon', \delta')$-DP, where
\[
\varepsilon' = \log(1 + q(e^{\varepsilon} - 1)), \quad \delta' = q \delta.
\]
\end{theorem}
When $\varepsilon$ is small enough, we have $\varepsilon' = O(q\varepsilon)$, indicating that subsampling amplifies privacy by a linear factor $q$. This implies that although subsampling may slightly increase the number of training steps, it effectively reduces the overall privacy loss. In other words, subsampling decreases the exposure probability of individual samples, significantly mitigating their influence on the model and thus providing stronger privacy protection.

Let $\mathcal{L}(\boldsymbol{\theta}) := \mathbb{E}_{\boldsymbol{x} \sim \mathcal{X}}[\ell(\boldsymbol{\theta}; \boldsymbol{x})]$ denote the expected risk, where $\ell(\boldsymbol{\theta}; \boldsymbol{x})$ represents the individual sample loss, and $\mathcal{X}$ is the underlying data distribution. We assume that $\mathcal{L}$ is $G$-Lipschitz smooth.

At iteration $t$, the gradient computed from a single sample is denoted as $\boldsymbol{g}_t(\boldsymbol{x}_i) := \nabla \ell(\boldsymbol{\theta}_t; \boldsymbol{x}_i)$, and the population gradient is defined as $\nabla \mathcal{L}(\boldsymbol{\theta}_t) := \mathbb{E}_{\boldsymbol{x} \sim \mathcal{X}}[\boldsymbol{g}_t(\boldsymbol{x})]$. The deviation from the expectation is denoted by $\boldsymbol{\xi}_t(\boldsymbol{x}_i) := \boldsymbol{g}_t(\boldsymbol{x}_i) - \nabla \mathcal{L}(\boldsymbol{\theta}_t)$, which is assumed to have zero mean.

For a mini-batch $\mathcal{B}_t$ of size $B$, let the raw gradient of sample $\boldsymbol{x}_i$ at step $t$ be denoted as $\boldsymbol{g}_t(\boldsymbol{x}_i) \in \mathbb{R}^d$. To reduce sensitivity and focus the privacy budget on the most informative dimensions, we first apply coordinate-wise sparsification using a predefined retention ratio $r \in (0,1)$. Let $\mathcal{I}_{\mathrm{sort}} = \{j_1, \dots, j_d\}$ denote the indices sorted in descending order of importance scores $s_j$, and define the retained coordinate set as
\[
\mathcal{I}_s = \{j_1, \dots, j_{\lfloor r \cdot d \rfloor} \}.
\]
The corresponding binary mask $\boldsymbol{m} \in \{0,1\}^d$ is defined by
\[
m_j = 
\begin{cases}
1, & \text{if } j \in \mathcal{I}_s, \\
0, & \text{otherwise}.
\end{cases}
\]

Each individual gradient is sparsified via element-wise multiplication:
\[
\boldsymbol{g}'_{t,i} := \boldsymbol{m} \odot \boldsymbol{g}_t(\boldsymbol{x}_i).
\]

Next, we apply $\ell_2$ norm clipping with threshold $C$ to the sparsified gradients:
\[
\bar{\boldsymbol{g}}_{t,i} := \frac{ \boldsymbol{g}'_{t,i} }{ \max\left(1, \frac{ \| \boldsymbol{g}'_{t,i} \|_2 }{C} \right) }.
\]

The mini-batch average of the clipped, sparsified gradients is then computed as:
\[
\bar{\boldsymbol{g}}_t := \frac{1}{B} \sum_{i \in \mathcal{B}_t} \bar{\boldsymbol{g}}_{t,i}.
\]
To preserve differential privacy, Gaussian noise is added only to the retained coordinates. Let $\boldsymbol{\zeta}_\sigma \sim \mathcal{N}(\boldsymbol{0}, \sigma^2 C^2 \boldsymbol{I}_d)$ be the isotropic noise vector. The final noisy update is computed as:
\[
\tilde{\boldsymbol{g}}_t := \bar{\boldsymbol{g}}_t + \boldsymbol{m} \odot \boldsymbol{\zeta}_\sigma.
\]

This formulation ensures that noise is injected only into selected coordinates while maintaining $(\varepsilon, \delta)$-differential privacy, and enables tighter control over both sensitivity and utility by concentrating noise on informative gradient dimensions.

Since coordinate-wise selection is performed using a fixed retention ratio \( r \in (0,1) \), and the binary mask \( \boldsymbol{m} \in \{0,1\}^d \) is constructed independent of the dataset, this selection does not increase sensitivity. Moreover, noise is added only to the retained coordinates. By the post-processing property of differential privacy, the overall mechanism remains differentially private.

\begin{theorem}[Privacy Guarantee of IGU]\label{the:Privacy Guarantee of DPIGU}
Let \( \boldsymbol{m} \in \{0,1\}^d \) be a fixed, data-independent binary mask corresponding to a retention ratio \( r = \frac{\|\boldsymbol{m}\|_0}{d} \).  
For each sample \( \boldsymbol{x}_i \), let the gradient at step \( t \) be \( \boldsymbol{g}_t(\boldsymbol{x}_i) \in \mathbb{R}^d \),  
and define the sparsified gradient as \( \boldsymbol{g}_t'(\boldsymbol{x}_i) := \boldsymbol{m} \odot \boldsymbol{g}_t(\boldsymbol{x}_i) \).  
Then the following mechanism:
\[
\tilde{\boldsymbol{g}}_t := \frac{1}{B} \sum_{i \in \mathcal{B}_t} 
\frac{\boldsymbol{g}_t'(\boldsymbol{x}_i)}{\max\left(1, \frac{\|\boldsymbol{g}_t'(\boldsymbol{x}_i)\|_2}{C} \right)} 
+ \boldsymbol{m} \odot \mathcal{N}(\boldsymbol{0}, \sigma^2 C^2 \boldsymbol{I}_d)
\]
satisfies \((\varepsilon, \delta)\)-differential privacy.
\end{theorem}

The proof is given in Appendix~\ref{proof:Privacy Guarantee of DPIGU}.

\begin{remark}
The retention ratio \( r \) determines the number of coordinates retained in the binary mask \( \boldsymbol{m} \). Since the mask is constructed in a data-independent way (e.g., via offline importance estimation), it does not introduce additional privacy loss. Furthermore, restricting noise injection to the retained coordinates improves the signal-to-noise ratio and enhances utility while maintaining rigorous privacy guarantees.
\end{remark}

\subsection{Convergence Analysis of Gradient Clipping in SGD}

Gradient clipping is a widely used technique in deep learning to prevent training instability caused by exploding gradients. In the context of differential privacy, clipping also plays a key role in bounding the sensitivity of gradient updates, thereby enabling privacy guarantees. Before analyzing differentially private optimizers, we first investigate the optimization behavior of SGD with clipped gradients.

The following theorem establishes a convergence upper bound on the inner product between true gradients and clipped stochastic gradients:

\begin{theorem}\label{the:clip+SGD}
Consider a $G$-smooth loss function $\mathcal{L}$. If SGD is performed with learning rate $\eta$, and the per-sample gradient is clipped at norm threshold $C$, then
\[
\frac{1}{T} \sum_{t=1}^{T} \mathbb{E} \left[\langle \nabla \mathcal{L}(\boldsymbol{\theta}_t), \bar{\boldsymbol{g}}_t \rangle \right] \leq \frac{1}{T} \eta \left(\mathbb{E}[\mathcal{L}(\boldsymbol{\theta}_1)] - \min_{\boldsymbol{\theta}} \mathcal{L}(\boldsymbol{\theta})\right) + \frac{\eta GC^2}{2T}.
\]
\end{theorem}

Here, $\bar{\boldsymbol{g}}_t$ denotes the average clipped gradient at step $t$. The clipping threshold $C$ governs the trade-off between gradient fidelity and sensitivity control. An excessively large $C$ may fail to suppress outlier gradients, while a small $C$ may overly constrain updates, impeding optimization. Therefore, $C$ should be chosen based on empirical gradient norms to balance stability and training effectiveness.

When differential privacy is imposed on SGD, the optimization dynamics are modified by adding calibrated Gaussian noise to clipped gradients, yielding a form of noisy SGD. This mechanism resembles stochastic processes used in sampling-based algorithms such as Langevin dynamics, and introduces a new source of variance that must be carefully analyzed for both privacy and convergence guarantees.

\subsection{Convergence Analysis of Importance-Based Gradient Updates with Gaussian Noise}
We begin by establishing a preliminary lemma on the energy retention property of top-$k$ sparsification:

\begin{lemma}
\label{lemma:topk_energy_retention}
Let $\boldsymbol{u} \in \mathbb{R}^d$ be an arbitrary vector, and let $\boldsymbol{m}^{(k)} \in \{0, 1\}^d$ be a deterministic top-$k$ mask such that $m_i^{(k)} = 1$ if $|u_i|$ is among the top-$k$ largest entries of $|\boldsymbol{u}|$, and $m_i^{(k)} = 0$ otherwise. Define the normalized energy retention ratio
\[
\alpha(\boldsymbol{u}) := \frac{\|\boldsymbol{m}^{(k)} \odot \boldsymbol{u}\|^2}{\|\boldsymbol{u}\|^2}.
\]
Then the following identity holds:
\[
\langle \boldsymbol{u}, \boldsymbol{m}^{(k)} \odot \boldsymbol{u} \rangle = \alpha(\boldsymbol{u}) \cdot \|\boldsymbol{u}\|^2.
\]
Furthermore, if $\boldsymbol{u}$ is non-negative and sorted as $u_{(1)} \leq \cdots \leq u_{(d)}$, then
\[
\alpha(\boldsymbol{u}) = \frac{\sum_{i=d-k+1}^{d} u_{(i)}^2}{\sum_{i=1}^{d} u_{(i)}^2}.
\]
In particular, if the magnitudes of the components are approximately uniform, we may expect $\alpha(\boldsymbol{u}) \approx \frac{k}{d}$.
\end{lemma}

The proof is given in~\ref{theproof:topk_energy_retention}.

\begin{corollary}
\label{cor:topk_mask_expectation}
Let $\boldsymbol{v} \in \mathbb{R}^d$ be a random vector with finite second moment, and let $\boldsymbol{m}^{(k)} \in \{0,1\}^d$ be a deterministic top-$k$ mask such that $m_i^{(k)} = 1$ if $|v_i|$ is among the top-$k$ largest coordinates of $|\boldsymbol{v}|$, and $m_i^{(k)} = 0$ otherwise. Then,
\[
\mathbb{E} \left\| \boldsymbol{m}^{(k)} \odot \boldsymbol{v} \right\|^2 
\leq \mathbb{E} \left[ \alpha(\boldsymbol{v}) \cdot \|\boldsymbol{v}\|^2 \right] 
\leq \alpha_t \cdot \mathbb{E} \|\boldsymbol{v}\|^2,
\]
where $\alpha(\boldsymbol{v}) := \dfrac{\|\boldsymbol{m}^{(k)} \odot \boldsymbol{v}\|^2}{\|\boldsymbol{v}\|^2}$ and $\alpha_t := \sup_{\boldsymbol{v}} \alpha(\boldsymbol{v}) \in (0,1)$ denotes the maximum energy retention ratio.
\end{corollary}

The proof is given in~\ref{theproof:topk_mask_expectation}.

\begin{theorem}\label{the:topk+noise+SGD}

Consider an optimization problem with a loss function $\mathcal{L}$ that is $G$-smooth. Let the gradient noise be $\boldsymbol{z}_\sigma \sim \mathcal{N}(\boldsymbol{0},  \sigma^2 \boldsymbol{I}_d)$, the learning rate be $\eta = \frac{1}{G\sqrt{T}}$, and the batch size be $B$. Suppose that the gradient bias is bounded by $\|\boldsymbol{z}_t\|^2 \leq \sigma_g^2$.
Assume that a fixed top-$k$ mask $\boldsymbol{m}^{(k)}$ is applied to the gradient at each iteration, and define the energy retention ratio as
\[
\alpha_t := \frac{\|\boldsymbol{m}^{(k)} \odot \nabla \mathcal{L}(\boldsymbol{\theta}_t)\|^2}{\|\nabla \mathcal{L}(\boldsymbol{\theta}_t)\|^2}.
\]
Then, the following holds
\[
\frac{1}{T} \sum_{t=1}^{T} \|\nabla \mathcal{L}(\boldsymbol{\theta}_t)\|^2 \leq \frac{2G}{\alpha_t\sqrt{T}} (\mathbb{E}[\mathcal{L}(\boldsymbol{\theta}_1)] - \min_{\boldsymbol{\theta}} \mathcal{L}(\boldsymbol{\theta})) + \frac{1}{\sqrt{T}} \left( \frac{\sigma_g^2}{B} + \frac{d\sigma^2}{B^2} \right).
\]
\end{theorem}
The proof is given in~\ref{proof:topk+noise+SGD}. This result suggests that top-$k$ sparsification introduces an additional factor $\alpha_t^{-1}$ to the convergence bound, reflecting the efficiency of information retention. As the retention ratio $r$ increases (i.e., more coordinates are preserved), $\alpha_t$ tends to increase, reducing the impact of sparsification. However, this may also increase the overall gradient magnitude and noise contribution, thus influencing training stability. Therefore, selecting an appropriate sparsity level involves a trade-off between privacy-induced noise reduction and optimization efficiency.

\subsection{Convergence Analysis of Importance-Based Updates in DPSGD}

In Theorem~\ref{the:topk+noise+SGD}, we focus on reducing the gradient noise introduced into the gradient update process and maintaining the differential privacy through the control of gradient noise. However, this method cannot directly control the gradient clipping in the case of differential privacy. In the next theorem , we introduce the gradient clipping mechanism to further increase the control of the gradient, reducing the extreme impact caused by the gradient noise.

\begin{theorem}\label{the:topk+clip+DPSGD}
Consider a DPSGD optimizer with per-sample gradient clipping $C$ and top-$k$ sparsification applied to each update. Assume the loss function $\mathcal{L}$ is $G$-smooth, the learning rate is $\eta$, and the batch size is $B$.
Let Gaussian noise $\boldsymbol{z}_{\mathrm{DP}} \sim \mathcal{N}(\boldsymbol{0}, \sigma^2 C^2 \boldsymbol{I}_d)$ be added to ensure $(\varepsilon, \delta)$-differential privacy. The noise and gradient are both masked by a fixed top-$k$ pattern, and let \(\alpha_t\) denote the energy retention ratio of the top-$k$ mask.

Then the following bound holds
\[
\mathbb{E}[\langle \nabla \mathcal{L}(\boldsymbol{\theta}_t), \tilde{\boldsymbol{g}}_t \rangle] 
\leq \frac{1}{\eta} \mathbb{E}[\mathcal{L}(\boldsymbol{\theta}_t) - \mathcal{L}(\boldsymbol{\theta}_{t+1})] 
+ \frac{G \eta}{2} \left( C^2 + \alpha_t \cdot \frac{d \sigma^2 C^2}{B} \right),
\]
where \(\tilde{\boldsymbol{g}}_t\) is the clipped, masked, and noise-injected update.
\end{theorem}

The proof is given in~\ref{proof:topk+clip+DPSGD}.
This result shows that incorporating both clipping and top-$k$ sparsification bounds the update error in expectation. When the gradient dimensionality $d$ increases, reducing the retention ratio $r$ can help lower computational cost and the effective noise dimension. However, decreasing $r$ also reduces the energy retention $\alpha_t$, potentially degrading convergence speed. In contrast, increasing the batch size $B$ improves the signal-to-noise ratio by averaging more per-sample gradients, thereby accelerating convergence and stabilizing updates.

\section{Experiments}
\label{sec:Experiments}
To demonstrate the utility of the proposed AdaDPIGU mechanism, we evaluate its performance on three standard benchmarks: MNIST~\cite{lecun1998gradient}, FashionMNIST~\cite{xiao2017fashion}, and CIFAR-10~\cite{krizhevsky2009learning}. These evaluations highlight the notable improvements achieved over DPSGD. The experiments were carried out using a PyTorch-based implementation on an NVIDIA GeForce RTX 4080 GPU. Several baseline models were also implemented for comparative analysis. Codes to reproduce our experiments are available at  \url{https://github.com/FangXieLab/AdaDPIGU}.

\subsection{Datasets and Network Structure}
We evaluate the effectiveness of differential privacy on three commonly used machine learning benchmark datasets: MNIST, FMNIST, and CIFAR-10. Although these datasets have been widely studied in the field of computer vision and are considered solved problems, achieving efficient model utility under strict privacy constraints remains a significant challenge.

MNIST consists of grayscale images representing handwritten digits, with 60{,}000 samples used for training and 10{,}000 for testing. Each image has a resolution of $28 \times 28$ pixels. The dataset includes ten digit classes, with each class containing around 7{,}000 samples. In a non-private scenario, traditional models with manually engineered features can reach an accuracy of 99.11\% after 20 epochs of training.

FashionMNIST (FMNIST) is designed as a drop-in replacement for MNIST, but with grayscale images of fashion items. It provides 60{,}000 examples for training and 10{,}000 for evaluation. Similar to MNIST, FMNIST contains ten distinct classes, each associated with $28 \times 28$ grayscale images of different product categories.

CIFAR-10 includes images from ten everyday object classes, offering 50{,}000 training samples and 10{,}000 for testing. Each image has dimensions $32 \times 32 \times 3$, corresponding to height, width, and RGB channels.

For the AdaDPIGU experiments conducted on the aforementioned datasets, the MNIST and FMNIST datasets follow the model architecture in Table~\ref{table:MNIST_and_fMNIST_model_architecture}, while CIFAR-10 utilizes the structure described in Table~\ref{table:CIFARt_model_architecture}. All datasets are trained using the categorical cross-entropy loss function.

\begin{table}[h]
\centering
\caption{Architecture used for MNIST and FMNIST classification}
\label{table:MNIST_and_fMNIST_model_architecture}
\begin{tabular}{lc}
\toprule
\textbf{Layer} & \textbf{Parameters} \\
\midrule
Convolution & 16 filters of $8 \times 8$, stride 2, padding 2 \\
Max-Pooling & $2 \times 2$, stride 1 \\
Convolution & 32 filters of $4 \times 4$, stride 2, padding 2 \\
Max-Pooling & $2 \times 2$, stride 1 \\
Fully Connected & 32 units \\
Softmax & 10 output classes \\
\bottomrule
\end{tabular}
\end{table}

\begin{table}[h]
\centering
\caption{CIFAR-10 convolutional model architecture}
\label{table:CIFARt_model_architecture}
\begin{tabular}{lc}
\toprule
\textbf{Layer} & \textbf{Parameters} \\
\midrule
Convolution & 16 filters of 3x3, strides 1, padding 1 \\
Max-Pooling & 2x2, stride 2 \\
Convolution & 16 filters of 3x3, strides 1, padding 1  \\
Max-Pooling & 2x2, stride 2\\
Convolution & 32 filters of 3x3, strides 1, padding 1  \\
Max-Pooling & 2x2, stride 2\\
Fully Connected & 128 units \\
Softmax & 10 units \\
\bottomrule
\end{tabular}
\end{table}

During the DPSGD training phase, we followed recommended configurations from prior studies~\cite{tramer2020differentially} for the three benchmark datasets. In particular, we varied the privacy budget $\varepsilon$ over the range of 2 to 12, while fixing $\delta = 10^{-5}$. The noise multiplier $\sigma$ was tuned accordingly for each $\varepsilon$ value, following the procedure outlined in~\cite{tramer2020differentially,wei2022dpis} (see Table~\ref{tab:noise_level}). This parameter tuning approach is routinely used in privacy-aware learning to navigate the balance between utility and privacy, without weakening the underlying privacy guarantees.

In the DPSGD stage, we utilized the optimal hyperparameters suggested in~\cite{tramer2020differentially} for the three image datasets. During the experiments, the privacy budget $\varepsilon$ was varied between 2 and 12, with $\delta$ fixed at $10^{-5}$. The noise multiplier $\sigma$ was specifically calibrated for each $\varepsilon$ value following the procedure outlined in~\cite{tramer2020differentially,wei2022dpis} (see Table~\ref{tab:noise_level}). This parameter tuning approach is widely adopted in privacy-preserving machine learning and does not introduce any additional privacy leakage.

\begin{table}[h]
\centering
\caption{Noise level \(\sigma\)  at different \(\varepsilon\) values}
\label{tab:noise_level}
\begin{tabular}{ccc}
\toprule
\textbf{Dataset} & \textbf{MNIST} & \textbf{CIFAR-10} \\
\midrule
$\varepsilon = 2$   & 4.64           & 3.62             \\
$\varepsilon = 4$   & 2.49           & 1.98             \\
$\varepsilon = 6$   & 1.79           & 1.45             \\
$\varepsilon = 8$   & 1.45           & 1.2              \\
$\varepsilon = 10$  & 1.25           & 1.05             \\
$\varepsilon = 12$  & 1.12           & 0.95             \\
\bottomrule
\end{tabular} 
\end{table}
A smaller privacy budget \( \varepsilon \) implies stronger privacy guarantees, as it requires injecting a higher magnitude of noise, thereby lowering the risk of information leakage. Similarly, a larger noise scale \( \sigma \) introduces more perturbation, which improves privacy at the potential cost of reduced training performance. As \( \varepsilon \) increases, the corresponding \( \sigma \) value decreases, since a looser privacy constraint permits injecting less noise, which in turn enhances model accuracy.

\subsection{Impact of Pruning Rate on Accuracy and Efficiency}
\begin{figure}[h]
    \centering
    \includegraphics[width=1.0\textwidth]{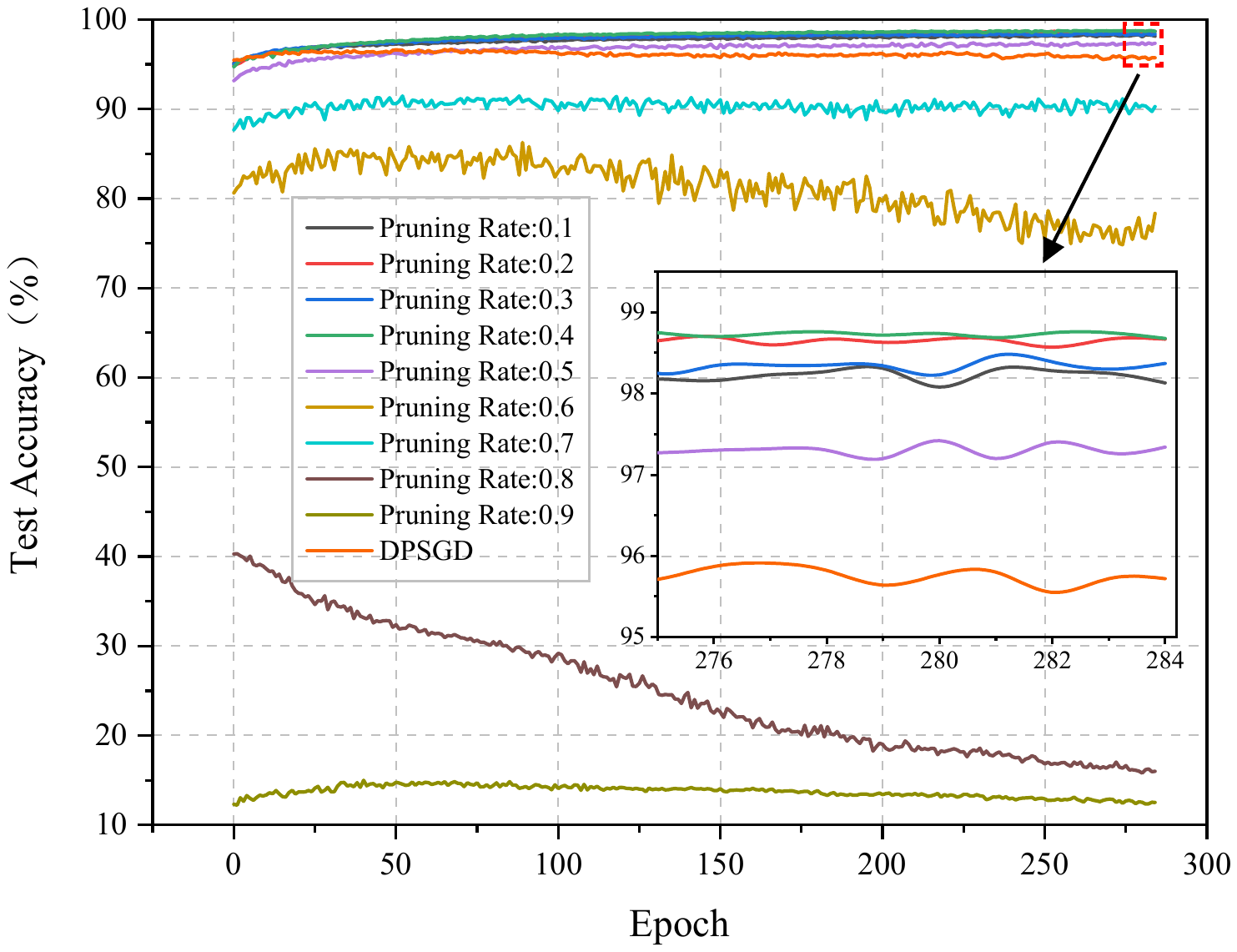}
    \caption{Test accuracy vs. epochs for different pruning rates}
    \label{fig:p-DPSGD}
\end{figure}

Figure~\ref{fig:p-DPSGD} illustrates the test accuracy trends on the MNIST dataset under different pruning configurations within the DPSGD setting. The results indicate that pruning significantly influences model accuracy, and this effect is strongly correlated with the resulting sparsity.

To determine the number of important gradients to retain, we define a sparsity retention ratio \(r \in (0,1)\), and compute \(k = \lfloor r \cdot d \rfloor\), where \(d\) is the total number of parameters. The top-\(k\) coordinates are selected based on their gradient importance scores accumulated during the DP pretraining phase. In this section, we empirically explore different pruning ratios corresponding to different \(k\) values, and observe that a pruning rate of 40\% (i.e., \(r = 0.6\)) achieves the best balance between accuracy and sparsity. This validates that the choice of \(k\) can be guided by tuning the pruning rate \(r\) and evaluating downstream performance.

When the pruning rate is within the range of 0.1 to 0.4, the test accuracy remains high and stable throughout training. In fact, these pruned models consistently outperform the baseline DPSGD model (i.e., without pruning), suggesting that moderate pruning may contribute to better generalization. This effect is especially prominent at pruning rate 0.4, which corresponds to retaining 60\% of the original parameters. As shown in the magnified inset of the figure, this setting achieves the highest accuracy in the late training stage, with minimal fluctuation across epochs.

We observe a clear ranking pattern in test accuracy for pruning rates between 0.1 and 0.4, specifically: $\mathrm{ACC}_{0.4} > \mathrm{ACC}_{0.2} > \mathrm{ACC}_{0.3} > \mathrm{ACC}_{0.1}$. This ordering indicates that 40\% pruning strikes a particularly favorable trade-off between model sparsity and representation capacity. We conjecture that such pruning effectively removes redundant or less informative parameters while preserving critical components of the network.

In contrast, higher pruning rates (e.g., above 0.5) result in significant degradation in test accuracy, especially for pruning rates 0.8 and 0.9. This suggests that aggressive pruning severely limits the model's expressiveness, making it difficult to recover useful representations under the added noise from differential privacy.

\subsection{Trade-off Between Privacy Budget and Model Utility}
\begin{figure}[h]
    \centering
    \includegraphics[width=\textwidth]{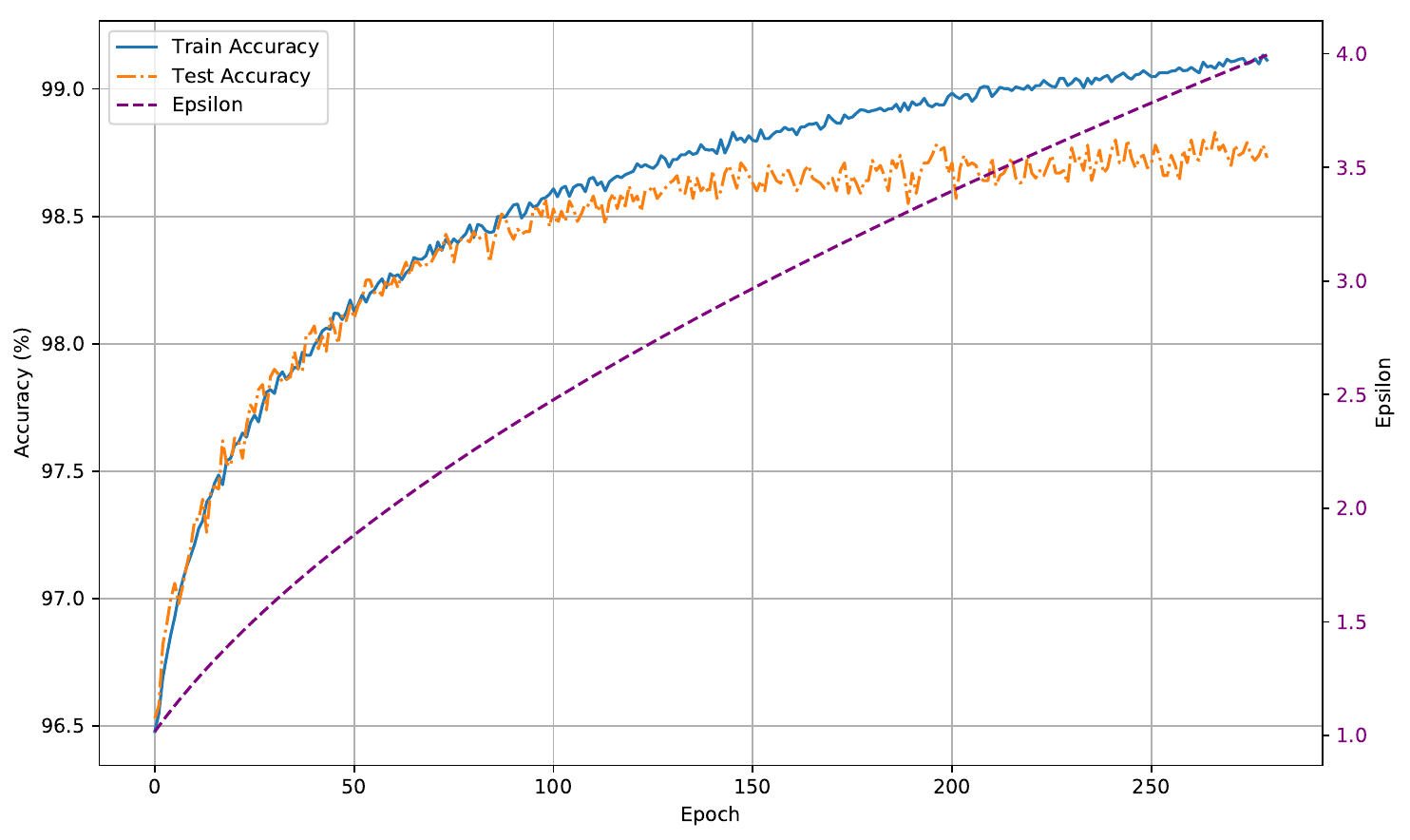}
    \caption{Training and test accuracy along with accumulated privacy budget over training epochs}
    \label{fig:lr0.2_stage3}
\end{figure}

Figure~\ref{fig:lr0.2_stage3} depicts the progression of training and testing performance along with the accumulation of the privacy budget ($\varepsilon$) during model optimization. As training proceeds over more epochs, both training and test accuracy exhibit steady improvement, while the total privacy cost increases correspondingly. This phenomenon highlights the inherent trade-off in differentially private learning: achieving higher model utility necessitates a greater expenditure of privacy budget over time.

At the end of training, with a total privacy budget of \(\varepsilon = 4\), our model achieves a test accuracy of 98.99\% on the MNIST dataset. This result demonstrates that high utility can still be maintained under a reasonably tight privacy constraint.

For reference, in the non-private setting, traditional hand-crafted feature models can reach up to 99.11\% test accuracy~\cite{tramer2020differentially}. Remarkably, our differentially private model, despite using only 60\% of the gradient updates via structured sparsification, achieves 98.99\% accuracy—nearly matching the non-private baseline. This result demonstrates the capability of the proposed approach to retain learning performance while substantially reducing the information disclosed in each update.

\begin{figure}[h]
    \centering
    \includegraphics[width=\textwidth]{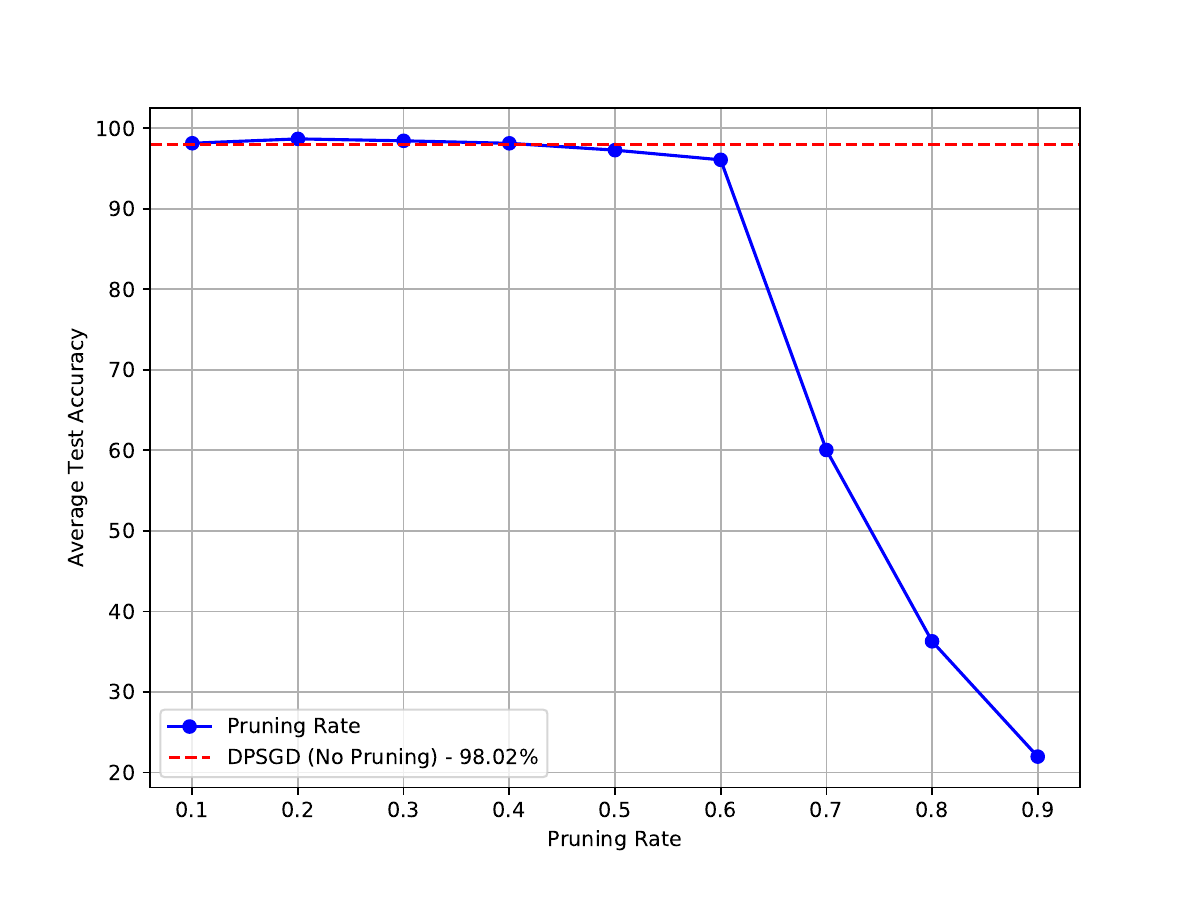}
    \caption{Test Accuracy vs. Pruning Rate (Batch Size 1000, Learning Rate 0.1)}
    \label{fig:epsilon4_pruning_accuracy_plot}
\end{figure}

Figure~\ref{fig:epsilon4_pruning_accuracy_plot} illustrates how test accuracy varies with different pruning ratios, under a constant experimental setup: batch size of 1000, learning rate of 0.1, and privacy parameter $\varepsilon = 4$. The blue curve shows the mean accuracy across the final 50 training epochs for each pruning level. The red dashed line marks the reference accuracy (98.02\%) obtained by applying DPSGD without pruning.

These observations suggest that pruning rates between 0.1 and 0.6 generally yield competitive or even superior performance compared to the unpruned baseline. The model maintains a comparable accuracy to the unpruned baseline, indicating that moderate pruning can effectively preserve essential information while maintaining generalization performance. In particular, pruning rates of 0.3 to 0.5 yield nearly indistinguishable performance compared to the full model, suggesting a degree of redundancy in the original parameter space.

However, when the pruning rate exceeds 0.6, model performance deteriorates rapidly. This significant decline in test accuracy highlights the detrimental effect of excessive sparsification, which likely removes critical parameters and impairs the model’s learning capacity under DP constraints.

\subsection{Privacy–Utility Cost Trade-off under Different Hyperparameters}
We conducted a systematic evaluation on the MNIST dataset by exploring various combinations of batch sizes (ranging from 500 to 1500) and learning rates (from 0.01 to 0.2). Our goal was to understand how these hyperparameters influence the trade-off between model utility, privacy protection, and computational cost under the DPSGD framework.

\begin{figure}[h]
    \centering
    \includegraphics[width=\textwidth]{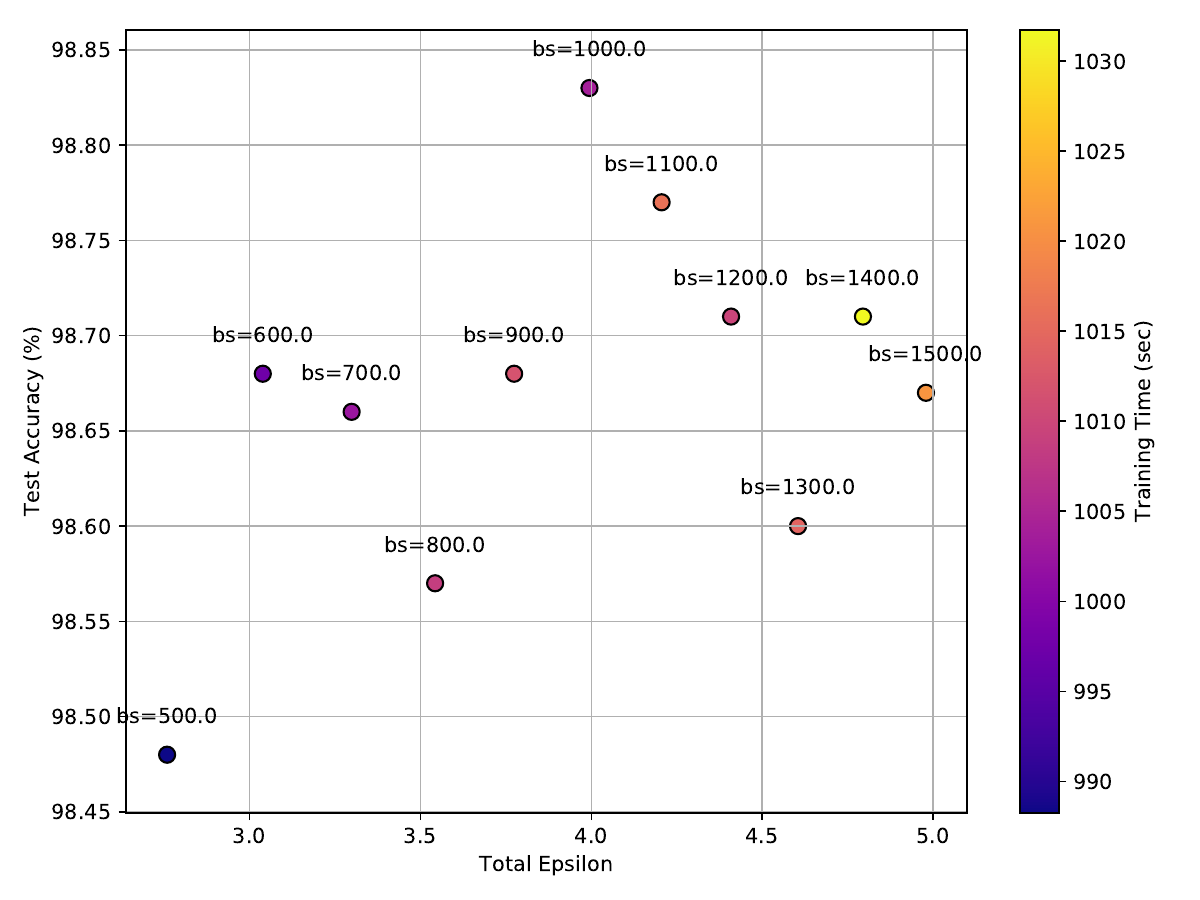}
    \caption{Visualization of the privacy–utility cost trade-off under AdaDPIGU on MNIST.}
    \label{fig:epsilon-batchsize-tradeoff}
\end{figure}

Figure~\ref{fig:epsilon-batchsize-tradeoff} visualizes the final test accuracy against the corresponding privacy budget \(\varepsilon\) for different batch sizes. Each point represents a particular configuration, where the color denotes the total training time (darker colors indicate longer durations), and the size of each marker corresponds to the total number of training steps—note that smaller batch sizes naturally require more iterations to complete a full epoch, hence resulting in larger markers.

The results show a clear trend: smaller batch sizes (e.g., 500–700) tend to yield lower privacy budgets, indicating stronger privacy guarantees. However, these configurations incur longer training times and higher computational overhead due to the increased number of gradient update steps. On the other hand, larger batch sizes (e.g., 1300–1500) reduce training time but result in larger final \(\varepsilon\) values, thus providing weaker privacy protection.

Notably, moderate batch sizes (around 900 to 1100) offer a well-balanced trade-off, achieving high test accuracy (above 98.75\%), reasonable training time, and a privacy budget typically within \(\varepsilon \approx 3.7\) to \(\varepsilon \approx 4.2\). These findings highlight the importance of hyperparameter tuning in differentially private training, and suggest that properly chosen batch sizes can lead to significant gains in both utility and efficiency without substantially compromising privacy.

\subsection{Comparison with State-of-the-Art Methods}
\begin{table}[h]
	\centering
	\caption{Results of classification accuracy}
	\label{table:classification_accuracy}
	\begin{tabularx}{\linewidth}{l l c c c}
		\toprule
		Dataset & Method & $\varepsilon=2$ & $\varepsilon=4$ & non-private \\
		\midrule
		MNIST 
		& AdaDPIGU & 98.06\% & \textbf{98.99\%} & \textbf{99.11\%} \\
		& DPSGD~\citep{abadi2016deep} & 95.10\% & 97.25\% & - \\
		& DPSUR~\citep{fu2023dpsur} & \textbf{98.70\%} & 98.95\% & - \\
		& DPSGD-HF~\citep{tramer2020differentially} & 98.39\% & 98.56\% & - \\
		& DPSGD-TS~\citep{papernot2021tempered} & 97.87\% & 98.51\% & - \\
		\addlinespace
		FMNIST 
		& AdaDPIGU & 86.93\% & \textbf{89.92\%} & \textbf{90.98\%} \\
		& DPSGD~\citep{abadi2016deep} & 79.89\% & 84.90\% & - \\
		& DPSUR~\citep{fu2023dpsur} & \textbf{89.34\%} & 90.18\% & - \\
		& DPSGD-HF~\citep{tramer2020differentially} & 87.96\% & 89.06\% & - \\
		& DPSGD-TS~\citep{papernot2021tempered} & 85.23\% & 86.92\% & - \\
		\addlinespace
		CIFAR-10 
		& AdaDPIGU & \textbf{70.90\%} & \textbf{73.21\%} & \textbf{71.12\%} \\
		& DPSGD~\citep{abadi2016deep} & 50.53\% & 58.19\% & - \\
		& DPSUR~\citep{fu2023dpsur} & 69.40\% & 71.45\% & - \\
		& DPSGD-HF~\citep{tramer2020differentially} & 66.32\% & 70.04\% & - \\
		& DPSGD-TS~\citep{papernot2021tempered} & 56.74\% & 60.96\% & - \\
		\bottomrule
	\end{tabularx}
\end{table}

We evaluate the performance of AdaDPIGU against four representative differentially private optimization methods (DPSGD, DPSUR, DPSGD-HF, and DPSGD-TS) on three benchmark datasets: MNIST, FMNIST, and CIFAR-10. Table~\ref{table:classification_accuracy} reports classification accuracy under privacy budgets $\varepsilon = 2$ and $\varepsilon = 4$, as well as non-private results for reference.

AdaDPIGU consistently achieves competitive or superior performance across datasets and privacy levels, demonstrating a favorable privacy–utility trade-off. On MNIST, AdaDPIGU reaches 98.99\% accuracy at $\varepsilon = 4$, closely approaching the non-private upper bound of 99.11\% and outperforming all other private baselines. In an extended experiment not shown in the table, AdaDPIGU further achieves 99.12\% at $\varepsilon = 8$ using only 60\% of the gradient coordinates, confirming the potential of sparsity to retain utility under relaxed privacy constraints.

Although the gains under tight privacy (e.g., $\varepsilon = 2$) are less pronounced on simpler datasets like MNIST and FMNIST, the advantages become more evident on complex tasks. In particular, on CIFAR-10, AdaDPIGU achieves 73.21\% accuracy at $\varepsilon = 4$, not only surpassing all other private methods but also exceeding the reported non-private baseline. This suggests that the structured gradient sparsification and adaptive clipping strategy in AdaDPIGU introduces a beneficial regularization effect, which helps mitigate overfitting and improves generalization, especially in high-dimensional and noisy settings.

In some cases, the improved accuracy is also attributed to AdaDPIGU's tighter clipping bound estimation, which reduces unnecessary noise injection and improves signal preservation under the same privacy constraint.

In summary, AdaDPIGU demonstrates strong adaptability across diverse tasks and privacy levels. Its robustness on both simple and challenging datasets highlights its potential as a practical and efficient solution for privacy-preserving deep learning.

\section{Conclusion}
In differentially private training, excessive noise injected into high-dimensional gradients often degrades model utility. However, many gradient coordinates are consistently negligible and contribute little to optimization. To address this issue, we propose AdaDPIGU, which enhances the DPSGD framework by combining gradient sparsification with coordinate-wise \emph{adaptive noise control}. In the pretraining phase, we accumulate gradient importance scores under differential privacy protection and retain only a fixed proportion of important coordinates for subsequent updates. In the training phase, AdaDPIGU applies a coordinate-wise adaptive clipping strategy that dynamically adjusts the noise scale for each coordinate based on its gradient statistics, thereby improving the signal-to-noise ratio of important updates while reducing noise on uninformative dimensions. Our theoretical analysis shows that AdaDPIGU satisfies $(\varepsilon, \delta)$-differential privacy and retains convergence guarantees.

Although AdaDPIGU demonstrates strong performance, it has several limitations. First, despite reducing noise from unimportant gradients, the method still requires careful allocation of the privacy budget. Under strict privacy constraints, balancing sparsification and noise injection may demand further tuning. Second, AdaDPIGU relies on a ranking-based heuristic to identify important gradients. In highly non-stationary or complex tasks, this approach may fail to capture the truly critical coordinates, resulting in suboptimal updates. Lastly, while the method performs well on standard datasets such as MNIST, its scalability to large-scale applications, especially in natural language processing or computer vision, remains to be validated.

Future research may explore dynamic privacy budget allocation strategies, where the budget is adaptively assigned based on the importance of parameters or gradients during training. This could improve both privacy efficiency and model performance in sensitive applications. Another promising direction is to scale AdaDPIGU to large models used in complex domains such as natural language processing (e.g., BERT, GPT) and computer vision (e.g., ResNet). Investigating the effectiveness of gradient sparsification and noise reduction in such settings is critical for practical deployment. Finally, a deeper exploration of the privacy-utility trade-off is essential. Although AdaDPIGU shows strong performance on benchmark datasets, further studies are needed to quantify its utility under varying privacy budgets and tasks. A detailed analysis will inform best practices for deploying AdaDPIGU in real-world privacy-preserving systems.


\acks{\noindent F. Xie was supported in part by the Guangdong Basic and Applied Basic Research Foundation (grant number 2023A1515110469), in part by the Guangdong Provincial Key Laboratory IRADS (grant number 2022B1212010006), and in part by the grant of Higher Education Enhancement Plan (2021-2025) of "Rushing to the Top, Making Up Shortcomings and Strengthening Special Features" (grant number 2022KQNCX100).}


\newpage

\appendix

\section{Preliminary on Differential Privacy}
\begin{definition}[R\'enyi Divergence]
\label{def:renyi_divergence}
Let \( X \) and \( Y \) be two discrete random variables over a common support of \( n \) elements, with probability mass functions \( \{p_i\} \) and \( \{q_i\} \), respectively. The R\'enyi divergence of order \( \alpha > 0 \), \( \alpha \neq 1 \), from \( X \) to \( Y \) is defined as
\[
D_\alpha(X \parallel Y) = \frac{1}{\alpha - 1} \log \left( \sum_{i=1}^{n} \frac{p_i^\alpha}{q_i^{\alpha - 1}} \right),
\]
where \( p_i = \Pr[X = x_i] \) and \( q_i = \Pr[Y = x_i] \).
\end{definition}

\begin{definition}[R\'enyi Divergence for Continuous Distributions]
\label{def:renyi_continuous}
Let \( p(x) \) and \( q(x) \) be two continuous probability density functions defined over a common domain \( \mathcal{X} \subseteq \mathbb{R}^d \). The R\'enyi divergence of order \( \alpha > 0 \), \( \alpha \neq 1 \), from \( p \) to \( q \) is defined as
\[
D_\alpha(p \parallel q) 
= \frac{1}{\alpha - 1} \log \int_{\mathcal{X}} p(x)^\alpha q(x)^{1 - \alpha} \, dx
= \frac{1}{\alpha - 1} \log \mathbb{E}_{x \sim q} \left( \frac{p(x)}{q(x)} \right)^\alpha,
\]
where \( \mathbb{E}_{x \sim q} \) denotes the expectation with respect to the distribution \( q(x) \).
\end{definition}

\begin{proposition}[RDP~\cite{mironov2017renyi}]\label{proposition:RDP+MGF}
For any $\alpha \geq 0$, if at time step $t$ the randomized mechanism $M_t$ satisfies
\[
\left(\alpha + 1, \frac{K_{M_t}(\alpha)}{\alpha}\right)\text{-RDP},
\]
then the composed mechanism $M$ over $K$ iterations satisfies at least
\[
\left(\alpha + 1, \frac{ \sum_{t=1}^K K_{M_t}(\alpha) }{\alpha}\right)\text{-RDP}.
\]
\end{proposition}

This proposition is based on the additivity of RDP, the Rényi divergence of several mechanisms satisfying RDP accumulates under composition.

\begin{corollary}[From RDP to DP]\label{corollary:From RDP to DP}
Combining Proposition~\ref{From RDP to DP} and Proposition~\ref{proposition:RDP+MGF}, we obtain the following result.
If the randomized training mechanism $M_t$ satisfies RDP at time step $t$, then
\[
\left(\frac{K_{M_t}(\alpha) + \log(1/\delta)}{\alpha}, \delta\right)\text{-DP},
\]
therefore, we can determine the minimal privacy loss $\varepsilon$ under a given $\delta$ via tuning the Rényi order
\[
\varepsilon(\delta) = \min_\alpha \frac{K_{M_t}(\alpha) + \log(1/\delta)}{\alpha}.
\]
Conversely, given a fixed $\varepsilon$, the minimal achievable $\delta$ is
\[
\delta = \min_\alpha \exp\left(K_{M_t}(\alpha) - \alpha \varepsilon\right).
\]
\end{corollary}

These results indicate that we can use the moment generating function (MGF) to optimize differential privacy accounting, and determine optimal privacy compositions by tuning the Rényi order $\alpha$.

\section{Deferred Proofs for Section~\ref{sec:Theoretical Analysis}}

\subsection{Proof of Theorem~\ref{the:Privacy Guarantee of DPIGU}}\label{proof:Privacy Guarantee of DPIGU}
\begin{proof}
We analyze the privacy guarantee of AdaDPIGU under the Gaussian mechanism with top-$k$ sparsification.

Let \(\boldsymbol{g}_t(\boldsymbol{x}_i)\) denote the clipped per-sample gradient at step \(t\). Given a clipping threshold \(C\), the \(\ell_2\)-sensitivity of the aggregated gradient between any pair of neighboring datasets \(D\) and \(D'\) (differing in at most one sample) satisfies
\[
\max_{D, D'} \left\| \sum_{\boldsymbol{x}_i \in D} \boldsymbol{g}_t(\boldsymbol{x}_i) - \sum_{\boldsymbol{x}_i \in D'} \boldsymbol{g}_t(\boldsymbol{x}_i) \right\|_2 \leq C.
\]

Now suppose a fixed top-\(k\) binary mask \(\boldsymbol{m} \in \{0, 1\}^d\) is applied to every gradient update. The masked per-sample gradient becomes
\[
\boldsymbol{g}_t'(\boldsymbol{x}_i) := \boldsymbol{x} \odot \boldsymbol{g}_t(\boldsymbol{x}_i).
\]

Since the mask is fixed and applied element-wise, it is a post-processing operation, which does not increase sensitivity. Therefore, the masked gradients satisfy the same sensitivity bound
\[
\max_{D, D'} \left\| \sum_{\boldsymbol{x}_i \in D} \boldsymbol{g}_t'(\boldsymbol{x}_i) - \sum_{\boldsymbol{x}_i \in D'} \boldsymbol{g}_t'(\boldsymbol{x}_i) \right\|_2 \leq C.
\]

In the Gaussian mechanism, the added noise before masking is drawn as
\[
\boldsymbol{\zeta}_t \sim \mathcal{N}(\boldsymbol{0}, \sigma^2 C^2 \boldsymbol{I}_d).
\]

Applying the same fixed top-\(k\) mask yields
\[
\boldsymbol{m} \odot \boldsymbol{\zeta}_t \sim \mathcal{N}\left(\boldsymbol{0}, \sigma^2 C^2 \operatorname{diag}(\boldsymbol{m})\right),
\]
where \(\operatorname{diag}(\boldsymbol{m})\) is a diagonal matrix that zeroes out all unmasked coordinates.
This masking operation restricts noise to a lower-dimensional subspace without affecting the statistical independence of the retained coordinates. Specifically
\[
\mathbb{E}[\boldsymbol{m} \odot \boldsymbol{\zeta}_t] = \boldsymbol{0}, \quad
\text{Cov}[\boldsymbol{m} \odot \boldsymbol{\zeta}_t] = \sigma^2 C^2 \operatorname{diag}(\boldsymbol{m}).
\]

Since the masking is deterministic and independent of the dataset, and the overall update process remains a valid Gaussian mechanism with the same sensitivity and scaled noise, the differential privacy guarantee is preserved.

By the post-processing theorem of differential privacy~\cite{dwork2014algorithmic}, the masked mechanism satisfies the same \((\varepsilon, \delta)\)-DP guarantee as the unmasked Gaussian mechanism.

\end{proof}

\subsection{Proof of Theorem~\ref{the:DPSGD}}\label{proof:DPSGD}
\begin{proof}
Assume the training mechanism at round \( t \) is denoted as \( M_t \), which maps neighboring datasets \( d, d' \in \mathcal{D} \) to output distributions over model parameters \( \boldsymbol{\theta}_t \). We consider the following condition on the output distribution similarity
\[
\sup_{\boldsymbol{\theta}_t \in \mathcal{R},\, d, d'} \left| \log \frac{\Pr[M_t(d) = \boldsymbol{\theta}_t]}{\Pr[M_t(d') = \boldsymbol{\theta}_t]} \right| \leq \varepsilon.
\]

To analyze the accumulation of privacy loss over multiple training steps, we follow the composition analysis introduced by Abadi et al.~\cite{abadi2016deep}. Let the model at round \( t \) receive parameters \( \boldsymbol{\theta}_{t-1} \) from the previous round, and update them using stochastic gradient descent.

We define the privacy loss random variable
\[
c(\boldsymbol{\theta}_t, M_t, \boldsymbol{\theta}_{t-1}, d, d') 
= \log \frac{\Pr[M_t(\boldsymbol{\theta}_{t-1}, d) = \boldsymbol{\theta}_t]}{\Pr[M_t(\boldsymbol{\theta}_{t-1}, d') = \boldsymbol{\theta}_t]}, 
\quad \boldsymbol{\theta}_t \sim M_t(\boldsymbol{\theta}_{t-1}, d).
\]

This random variable measures the change in output distribution of the mechanism when applied to neighboring datasets. To quantify the tail of this change, we consider the moment generating function (MGF) of the privacy loss, defined for Rényi order \( \alpha \) as
\begin{align*}
K_{M_t}^{d,d'}(\alpha) 
&:= \log \mathbb{E}_{\boldsymbol{\theta}_t \sim M_t(\boldsymbol{\theta}_{t-1}, d)} \left[ \exp\left( \alpha \cdot c(\boldsymbol{\theta}_t, M_t, \boldsymbol{\theta}_{t-1}, d, d') \right) \right] \\
&= \log \mathbb{E}_{\boldsymbol{\theta}_t \sim M_t(\boldsymbol{\theta}_{t-1}, d)} \left[ \left( \frac{\Pr[M_t(\boldsymbol{\theta}_{t-1}, d) = \boldsymbol{\theta}_t]}{\Pr[M_t(\boldsymbol{\theta}_{t-1}, d') = \boldsymbol{\theta}_t]} \right)^\alpha \right].
\end{align*}

By a change of measure, this expression can be rewritten as
\[
K_{M_t}^{d,d'}(\alpha) 
= \log \mathbb{E}_{\boldsymbol{\theta}_t \sim M_t(\boldsymbol{\theta}_{t-1}, d')} \left[ \left( \frac{\Pr[M_t(\boldsymbol{\theta}_{t-1}, d)]}{\Pr[M_t(\boldsymbol{\theta}_{t-1}, d')]} \right)^{\alpha + 1} \right].
\]

Since the training process consists of \( K \) rounds of mechanisms \( M_1, \cdots, M_K \), where each round depends only on the previous round's model parameter, the training procedure forms a Markov chain. This Markov property implies that the privacy loss random variables at different steps are conditionally independent
\[
\Pr[M_t(\boldsymbol{\theta}_{t-1}, d) = \boldsymbol{\theta}_t \mid M_{t-1}(\boldsymbol{\theta}_{t-2}, d) = \boldsymbol{\theta}_{t-1}] = \Pr[M_t(\boldsymbol{\theta}_{t-1}, d) = \boldsymbol{\theta}_t].
\]

As a result, the total privacy loss across the training procedure \( M_{1:K} \) can be expressed as a sum of per-step losses
\[
c(\boldsymbol{\theta}_{1:K}, M_{1:K}, \boldsymbol{\theta}_0: \boldsymbol{\theta}_{K-1}, d, d') 
= \sum_{t=1}^K \log \frac{\Pr[M_t(\boldsymbol{\theta}_{t-1}, d) = \boldsymbol{\theta}_t]}{\Pr[M_t(\boldsymbol{\theta}_{t-1}, d') = \boldsymbol{\theta}_t]} 
= \sum_{t=1}^K c(\boldsymbol{\theta}_t, M_t, \boldsymbol{\theta}_{t-1}, d, d').
\]

Accordingly, the moment generating function (MGF) of the total privacy loss also satisfies the additive property
\[
K_{M_{1:K}}^{d,d'}(\alpha) = \sum_{t=1}^{K} K_{M_t}^{d,d'}(\alpha).
\]

To ensure a worst-case guarantee over all neighboring datasets \( d, d' \in \mathcal{D} \), we define the global MGF upper bound as
\[
K_{M_{1:K}}(\alpha) := \sup_{d, d' \in \mathcal{D}} K_{M_{1:K}}^{d,d'}(\alpha) = \sum_{t=1}^{K} \sup_{d, d' \in \mathcal{D}} K_{M_t}^{d,d'}(\alpha) = \sum_{t=1}^{K} K_{M_t}(\alpha).
\]

This shows that even in the worst-case scenario, the overall privacy loss of the composed mechanism remains additive in terms of Rényi divergence.

According to Definition~\ref{def:renyi_continuous}, we can directly relate the moment generating function $K_{M_t}^{d,d'}(\alpha)$ to the c divergence as follows
\begin{align*}
K_{M_t}^{d,d'}(\alpha) &= \log \mathbb{E}_{\boldsymbol{\theta}_t \sim M_t(\boldsymbol{\theta}_{t-1}, d)} \left[ \left( \frac{\Pr[M_t(\boldsymbol{\theta}_{t-1}, d) = \boldsymbol{\theta}_t]}{\Pr[M_t(\boldsymbol{\theta}_{t-1}, d') = \boldsymbol{\theta}_t]} \right)^{\alpha} \right] \\
&= \alpha D_{\alpha+1} \left( M_t(\boldsymbol{\theta}_{t-1}, d) \, \| \, M_t(\boldsymbol{\theta}_{t-1}, d') \right).
\end{align*}

Therefore, the MGF provides a way to characterize the deviation between outputs of training mechanisms on neighboring datasets. Furthermore, we can use MGF to express the privacy guarantee parameter of RDP.
According to Corollary~\ref{corollary:From RDP to DP}, we analyze differential privacy from a probabilistic perspective. For any pair of neighboring datasets \( d, d' \), if the single-round privacy loss is defined as \( c(\boldsymbol{\theta}_t, M_t, \boldsymbol{\theta}_{t-1}, d, d') \), then the cumulative privacy loss over \( K \) steps is
\[
c(\boldsymbol{\theta}_t, M) = \sum_{t=1}^K c(\boldsymbol{\theta}_t, M_t, \boldsymbol{\theta}_{t-1}, d, d').
\]

To bound the tail of this loss, we apply Markov’s inequality to the moment generating function of the cumulative privacy loss
\begin{align*}
\Pr[c(\boldsymbol{\theta}_t, M) > \varepsilon] 
&= \Pr[\exp(\alpha c(\boldsymbol{\theta}_t, M)) > \exp(\alpha \varepsilon)] \\
&\leq \frac{\mathbb{E}[\exp(\alpha c(\boldsymbol{\theta}_t, M))]}{\exp(\alpha \varepsilon)}.
\end{align*}

Assuming the mechanisms \( M_1, \dots, M_K \) are conditionally independent given the Markov assumption, the expectation decomposes as
\[
\mathbb{E}[\exp(\alpha c(\boldsymbol{\theta}_t, M))] 
= \prod_{t=1}^{K} \mathbb{E}[\exp(\alpha c(\boldsymbol{\theta}_t, M_t))] 
= \exp\left( \sum_{t=1}^K K_{M_t}^{d,d'}(\alpha) \right).
\]

Therefore, we obtain the following bound
\[
\Pr[c(\boldsymbol{\theta}_t, M) > \varepsilon] 
\leq \exp\left( \sum_{t=1}^K K_{M_t}^{d,d'}(\alpha) - \alpha \varepsilon \right) 
= \delta.
\]

Rearranging, we find that to ensure \((\varepsilon, \delta)\)-differential privacy, it suffices to guarantee
\[
\sum_{t=1}^{K} K_{M_t}^{d,d'}(\alpha) \leq \alpha \varepsilon + \log \left( \frac{1}{\delta} \right).
\]

As stated earlier, although overlapping between datasets may occur, the worst-case privacy loss only arises under extreme differences between neighboring datasets. To ensure rigorous privacy guarantees, we must consider the maximum Rényi divergence over all neighboring dataset pairs \( (d, d') \). In practice, we directly compute the upper bound of the moment generating function (MGF), \( K_{M_t}^{d,d'}(\alpha) \), at each training step.

According to the analysis of Abadi et al.~\cite{abadi2016deep}, under the Gaussian mechanism with subsampling rate \( q \), the Rényi divergence at order \( \alpha \) satisfies
\[
K_{M_t}^{d,d'}(\alpha) \leq \frac{q^2 \alpha^2}{\sigma^2}.
\]

Assuming a total of \( T \) training steps with independent mechanisms, the MGF of the composed mechanism satisfies the additive property
\[
K_{M}(\alpha) = \sum_{t=1}^T K_{M_t}^{d,d'}(\alpha) \leq \frac{T q^2 \alpha^2}{\sigma^2}.
\]

Combining Corollary~\ref{corollary:From RDP to DP}, we can translate this Rényi bound into an \((\varepsilon, \delta)\)-DP guarantee by minimizing
\[
\delta = \min_\alpha \exp\left( K_M(\alpha) - \alpha \varepsilon \right).
\]

Substituting the bound into the expression gives
\[
\delta \leq \exp\left( \frac{T q^2 \alpha^2}{\sigma^2} - \alpha \varepsilon \right).
\]

Taking derivative with respect to \( \alpha \) and solving for the optimal order yields
\[
\alpha^* = \frac{\sigma^2 \varepsilon}{2 T q^2}.
\]

Plugging back gives the minimal achievable \( \delta \)
\[
\delta = \exp\left( -\frac{\sigma^2 \varepsilon^2}{4 T q^2} \right).
\]

Rearranging yields the required noise scale to ensure \((\varepsilon, \delta)\)-DP
\[
\sigma = \frac{2q \sqrt{T \log(1/\delta)}}{\varepsilon}.
\]
\end{proof}

\subsection{Proof of Theorem~\ref{the:clip+SGD}}\label{theproof:clip+SGD}

\begin{proof}
We start from the $G$-smoothness of the objective to derive a descent inequality involving the true gradient and the noisy, clipped update $\bar{\boldsymbol{g}}_t$. By taking expectations and bounding the gradient norm using the clipping threshold $C$, we obtain an upper bound on the expected inner product $\mathbb{E}[\langle \nabla \mathcal{L}(\boldsymbol{\theta}_t), \bar{\boldsymbol{g}}_t \rangle]$. Averaging over $T$ steps and applying the lower bound on the loss function, we arrive at a convergence-type bound that quantifies optimization progress under differentially private updates.

According to the $G$-smoothness assumption, the loss function satisfies
\begin{align*}
\mathcal{L}(\boldsymbol{\theta}_{t+1}) 
&\leq \mathcal{L}(\boldsymbol{\theta}_t) + \langle \nabla \mathcal{L}(\boldsymbol{\theta}_t), \boldsymbol{\theta}_{t+1} - \boldsymbol{\theta}_t \rangle + \frac{G}{2} \|\boldsymbol{\theta}_{t+1} - \boldsymbol{\theta}_t\|^2 \notag \\
&= \mathcal{L}(\boldsymbol{\theta}_t) - \eta \langle \nabla \mathcal{L}(\boldsymbol{\theta}_t), \bar{\boldsymbol{g}}_t \rangle + \frac{G \eta^2}{2} \|\bar{\boldsymbol{g}}_t\|^2.
\end{align*}
Taking expectation on both sides and rearranging terms yields
\begin{align*}
\mathbb{E}[\langle \nabla \mathcal{L}(\boldsymbol{\theta}_t), \bar{\boldsymbol{g}}_t \rangle] 
&\leq \frac{1}{\eta} \mathbb{E}\left[\mathcal{L}(\boldsymbol{\theta}_t) - \mathcal{L}(\boldsymbol{\theta}_{t+1})\right] 
+ \frac{G \eta}{2} \mathbb{E}[\|\bar{\boldsymbol{g}}_t\|^2] \\
&\leq \frac{1}{\eta} \mathbb{E}\left[\mathcal{L}(\boldsymbol{\theta}_t) - \mathcal{L}(\boldsymbol{\theta}_{t+1})\right] 
+ \frac{\eta G C^2}{2}.
\end{align*}
where the last inequality follows from the clipping bound $\|\bar{\boldsymbol{g}}_t\| \leq C$.
Taking average over $T$ steps
\[
\frac{1}{T} \sum_{t=1}^{T} \mathbb{E}[\langle \nabla \mathcal{L}(\boldsymbol{\theta}_t), \bar{\boldsymbol{g}}_t \rangle] 
\leq \frac{1}{T \eta} \mathbb{E}[\mathcal{L}(\boldsymbol{\theta}_1) - \mathcal{L}(\boldsymbol{\theta}_{t+1})] + \frac{\eta G C^2}{2T}.
\]
Since $
\mathbb{E}[\mathcal{L}(\boldsymbol{\theta}_{t+1})] \geq \min_\theta \mathcal{L}(\boldsymbol{\theta}),
$
we have
\[
\mathbb{E}[\mathcal{L}(\boldsymbol{\theta}_1) - \mathcal{L}(\boldsymbol{\theta}_{t+1})] 
= \mathbb{E}[\mathcal{L}(\boldsymbol{\theta}_1)] - \mathbb{E}[\mathcal{L}(\boldsymbol{\theta}_{t+1})] 
\leq \mathbb{E}[\mathcal{L}(\boldsymbol{\theta}_1)] - \min_\theta \mathcal{L}(\boldsymbol{\theta}).
\]
Thus, we can obtain
\[
\frac{1}{T} \sum_{t=1}^{T} \mathbb{E}[\langle \nabla \mathcal{L}(\boldsymbol{\theta}_t), \bar{\boldsymbol{g}}_t \rangle] 
\leq \frac{1}{T \eta} \left( \mathbb{E}[\mathcal{L}(\boldsymbol{\theta}_1)] - \min_\theta \mathcal{L}(\boldsymbol{\theta}) \right) + \frac{\eta G C^2}{2T}.
\]
\end{proof}

\subsection{Proof of Lemma~\ref{lemma:topk_energy_retention}}\label{theproof:topk_energy_retention}

\begin{proof}
The argument is adapted from~\cite{beznosikov2023biased}, which analyzes the energy preserved by top-$k$ projections.
By definition of the top-$k$ mask \( \boldsymbol{m}^{(k)} \), we have
\[
\langle \boldsymbol{u}, \boldsymbol{m}^{(k)} \odot \boldsymbol{u} \rangle 
= \sum_{i=1}^{d} m_i^{(k)} u_i^2 
= \|\boldsymbol{m}^{(k)} \odot \boldsymbol{u}\|^2.
\]

Thus, the energy retention ratio is given by
\[
\alpha(\boldsymbol{u}) := \frac{\|\boldsymbol{m}^{(k)} \odot \boldsymbol{u}\|^2}{\|\boldsymbol{u}\|^2}.
\]

To establish a lower bound on \(\alpha(\boldsymbol{u})\), assume that \(\boldsymbol{u}\) is non-negative and sorted in non-decreasing order: \( u_{(1)} \leq \cdots \leq u_{(d)} \). Then the top-$k$ mask selects the $k$ largest coordinates \( u_{(d-k+1)}, \dots, u_{(d)} \), and we obtain
\[
\alpha(\boldsymbol{u}) = \frac{\sum_{i=d-k+1}^{d} u_{(i)}^2}{\sum_{i=1}^{d} u_{(i)}^2}.
\]

Since both the numerator and denominator consist of squared non-negative terms and the numerator selects the largest \(k\) values, the minimum value of this ratio is achieved when all entries are equal. In this case, each \(u_i^2 = c\) for some constant \(c > 0\), and
\[
\alpha(\boldsymbol{u}) = \frac{k c}{d c} = \frac{k}{d}.
\]

Therefore, we conclude that
\[
\alpha(\boldsymbol{u}) \geq \frac{k}{d}.
\]
\end{proof}

\subsection{Proof of Corollary~\ref{cor:topk_mask_expectation}}\label{theproof:topk_mask_expectation}
\begin{proof}
By Lemma~\ref{lemma:topk_energy_retention}, for any realization of the random vector $\boldsymbol{v} \in \mathbb{R}^d$, we have
\[
\|\boldsymbol{m}^{(k)} \odot \boldsymbol{v}\|^2 = \alpha(\boldsymbol{v}) \cdot \|\boldsymbol{v}\|^2,
\]
taking expectation on both sides gives
\[
\mathbb{E} \|\boldsymbol{m}^{(k)} \odot \boldsymbol{v}\|^2 = \mathbb{E}[\alpha(\boldsymbol{v}) \cdot \|\boldsymbol{v}\|^2].
\]

Since $\alpha(\boldsymbol{v}) \leq \alpha_t := \sup_{\boldsymbol{v}} \alpha(\boldsymbol{v})$ by definition, and assuming $\boldsymbol{v}$ has finite second moment, we apply the inequality
\[
\mathbb{E}[\alpha(\boldsymbol{v}) \cdot \|\boldsymbol{v}\|^2] 
\leq \alpha_t \cdot \mathbb{E} \|\boldsymbol{v}\|^2.
\]

Therefore,
\[
\mathbb{E} \|\boldsymbol{m}^{(k)} \odot \boldsymbol{v}\|^2 
\leq \alpha_t \cdot \mathbb{E} \|\boldsymbol{v}\|^2,
\]
which completes the proof.
\end{proof}

\subsection{Proof of Theorem~\ref{the:topk+noise+SGD}}\label{proof:topk+noise+SGD}

\begin{proof}
The goal is to upper bound the average squared gradient norm under differentially private updates with sparsified gradients. The proof starts from the $G$-smoothness inequality to relate the loss decrease to the update direction. We substitute the masked and noise-perturbed gradient into the update rule and take expectations to isolate the alignment term between the true gradient and the update. Using Lemma~\ref{lemma:topk_energy_retention}, we quantify the energy retained by the top-$k$ mask via a factor $\alpha_t$. The norm of the masked noisy gradient is then bounded via Corollary~\ref{cor:topk_mask_expectation}, leading to an upper bound that accounts for both gradient variance and additive Gaussian noise. After rearranging terms and averaging over $T$ steps, we arrive at a convergence-style result that decays as $\mathcal{O}(1/\sqrt{T})$, with a dependence on $\alpha_t$, $B$, $d$, and $\sigma^2$.
From the $G$-Lipschitz smoothness condition, we have
\[
\langle \nabla f(\boldsymbol{y}), (\boldsymbol{x}-\boldsymbol{y}) \rangle = \nabla f(\boldsymbol{y})^T (\boldsymbol{x}-\boldsymbol{y}),
\]
\[
f(\boldsymbol{x}) \leq f(\boldsymbol{y}) + \nabla f(\boldsymbol{y})^T(\boldsymbol{x}-\boldsymbol{y}) + \frac{G}{2} \|\boldsymbol{x}-\boldsymbol{y}\|^2.
\]

Note that Gaussian noise is added to the aggregated gradient, and is scaled by the batch size $B$ before updating the model. Then we have
\[
\mathcal{L}(\boldsymbol{\theta}_{t+1}) \leq \mathcal{L}(\boldsymbol{\theta}_t) + \langle \nabla \mathcal{L}(\boldsymbol{\theta}_t), (\boldsymbol{\theta}_{t+1} - \boldsymbol{\theta}_t) \rangle + \frac{G}{2} \|\boldsymbol{\theta}_{t+1} - \boldsymbol{\theta}_t\|^2.
\]

The noise-injected SGD update rule is
\[
\boldsymbol{\theta}_{t+1} = \boldsymbol{\theta}_t - \eta 
\Bigg( \boldsymbol{m} \odot 
\bigg( \nabla \mathcal{L}(\boldsymbol{\theta}_t) + 
\frac{1}{B} \Big( \sum_i \boldsymbol{z}_{t,i} + \boldsymbol{z}_\sigma \Big) 
\bigg) \Bigg).
\]

Then we obtain
\[
\mathcal{L}(\boldsymbol{\theta}_{t+1}) 
\leq \mathcal{L}(\boldsymbol{\theta}_t) 
- \eta \Bigg\langle 
\nabla \mathcal{L}(\boldsymbol{\theta}_t),\ 
\boldsymbol{m} \odot 
\bigg( \nabla \mathcal{L}(\boldsymbol{\theta}_t) 
+ \frac{1}{B} \Big( \sum_i \boldsymbol{z}_{t,i} + \boldsymbol{z}_\sigma \Big) 
\bigg) 
\Bigg\rangle
\]

\[
+ \frac{\eta^2 G}{2} 
\Bigg\| 
\boldsymbol{m} \odot 
\bigg( \nabla \mathcal{L}(\boldsymbol{\theta}_t) 
+ \frac{1}{B} \Big( \sum_i \boldsymbol{z}_{t,i} + \boldsymbol{z}_\sigma \Big) 
\bigg) 
\Bigg\|^2.
\]

Taking expectation and rearranging, we have
\begin{align}
&\eta \, \mathbb{E} \left\langle 
\nabla \mathcal{L}(\boldsymbol{\theta}_t),\ 
\boldsymbol{m} \odot \left( 
\nabla \mathcal{L}(\boldsymbol{\theta}_t) + \frac{1}{B} \sum_i \boldsymbol{z}_{t,i} + \frac{1}{B} \boldsymbol{z}_\sigma 
\right) 
\right\rangle \notag \\
&\leq \mathbb{E}[\mathcal{L}(\boldsymbol{\theta}_t) - \mathcal{L}(\boldsymbol{\theta}_{t+1})] 
+ \eta^2 \mathbb{E} \left[ 
\frac{G}{2} \left\| 
\boldsymbol{m} \odot \left( 
\nabla \mathcal{L}(\boldsymbol{\theta}_t) + \frac{1}{B} \sum_i \boldsymbol{z}_{t,i} + \frac{1}{B} \boldsymbol{z}_\sigma 
\right) 
\right\|^2 
\right]
\label{eq:expected_inner_bound}
\end{align}

Since (by Lemma~\ref{lemma:topk_energy_retention})
\[
\langle \boldsymbol{u}, \boldsymbol{m}^{(k)} \odot \boldsymbol{u} \rangle = \alpha_t \|\boldsymbol{u}\|^2,
\]
where $\alpha_t := \|\boldsymbol{m}^{(k)} \odot \nabla \mathcal{L}(\boldsymbol{\theta}_t)\|^2 / \|\nabla \mathcal{L}(\boldsymbol{\theta}_t)\|^2$,
substituting back, we obtain
\begin{align*}
&\eta \, \mathbb{E} \left\langle 
\nabla \mathcal{L}(\boldsymbol{\theta}_t),\ 
\boldsymbol{m} \odot \left( \nabla \mathcal{L}(\boldsymbol{\theta}_t) + \frac{1}{B} \sum_i \boldsymbol{z}_{t,i} + \boldsymbol{z}_\sigma \right) 
\right\rangle \\
&= \eta \left\{
\mathbb{E} \langle \nabla \mathcal{L}(\boldsymbol{\theta}_t),\ \boldsymbol{m} \odot \nabla \mathcal{L}(\boldsymbol{\theta}_t) \rangle 
+ \mathbb{E} \left\langle \nabla \mathcal{L}(\boldsymbol{\theta}_t),\ 
\boldsymbol{m} \odot \left( \frac{1}{B} \sum_i \boldsymbol{z}_{t,i} + \boldsymbol{z}_\sigma \right) \right\rangle
\right\} \\
&= \eta \left\{ 
\alpha_t \|\nabla \mathcal{L}(\boldsymbol{\theta}_t)\|^2 
+ \alpha_t \, \mathbb{E} \left\langle \boldsymbol{g}_\theta,\ \frac{1}{B} \sum_i \boldsymbol{z}_{t,i} + \boldsymbol{z}_\sigma \right\rangle 
\right\} \\
&= \eta \alpha_t \|\nabla \mathcal{L}(\boldsymbol{\theta}_t)\|^2.
\end{align*}

By substituting into Equation~\ref{eq:expected_inner_bound}, we obtain
\begin{equation}
\eta \alpha_t \|\nabla \mathcal{L}(\boldsymbol{\theta}_t)\|^2 
\leq \mathbb{E}[\mathcal{L}(\boldsymbol{\theta}_t) - \mathcal{L}(\boldsymbol{\theta}_{t+1})] 
+ \frac{G \eta^2}{2} \mathbb{E} \left\| 
\boldsymbol{m} \odot \left( \nabla \mathcal{L}(\boldsymbol{\theta}_t) + \frac{1}{B} \sum_i \boldsymbol{z}_{t,i} + \boldsymbol{z}_\sigma \right) 
\right\|^2
\label{eq:final_bound}
\end{equation}

By expanding the squared norm term
\[
\frac{G \eta^2}{2} \, \mathbb{E} \left( \boldsymbol{m}^2 \left[ \left( \nabla \mathcal{L}(\boldsymbol{\theta}_t) \right)^2 
+ \frac{1}{B^2} \left( \sum_i \boldsymbol{z}_{t,i} + \boldsymbol{z}_\sigma \right)^2 
+ 2 \nabla \mathcal{L}(\boldsymbol{\theta}_t) \cdot \frac{1}{B} \left( \sum_i \boldsymbol{z}_{t,i} + \boldsymbol{z}_\sigma \right)
\right] \right)
\]
\[
= \mathbb{E} \left\| 
\boldsymbol{m} \odot \left( \nabla \mathcal{L}(\boldsymbol{\theta}_t) + \frac{1}{B} \sum_i \boldsymbol{z}_{t,i} + \boldsymbol{z}_\sigma \right) 
\right\|^2,
\]
where the expectation of the inner product between the gradient and the noise is zero, due to their independence.

Since (by Corollary~\ref{cor:topk_mask_expectation})
\begin{align}
&\mathbb{E} \left\| 
\boldsymbol{m} \odot \left( \nabla \mathcal{L}(\boldsymbol{\theta}_t) + \frac{1}{B} \sum_i \boldsymbol{z}_{t,i} + \boldsymbol{z}_\sigma \right) 
\right\|^2 \notag \\
&\leq \alpha_t \,
\mathbb{E} \left\| \nabla \mathcal{L}(\boldsymbol{\theta}_t) + \frac{1}{B} \left( \sum_i \boldsymbol{z}_{t,i} + \boldsymbol{z}_\sigma \right) \right\|^2.
\label{eq:mask_bound_alpha}
\end{align}

To further simplify the upper bound in Equation~\eqref{eq:mask_bound_alpha}, we expand the squared norm of the noisy gradient as
\begin{align*}
&\mathbb{E} \left\| \nabla \mathcal{L}(\boldsymbol{\theta}_t) + \frac{1}{B} \left( \sum_i \boldsymbol{z}_{t,i} + \boldsymbol{z}_\sigma \right) \right\|^2 \\
&= \mathbb{E} \left\| \nabla \mathcal{L}(\boldsymbol{\theta}_t) \right\|^2 
+ \mathbb{E} \left\| \frac{1}{B} \left( \sum_i \boldsymbol{z}_{t,i} + \boldsymbol{z}_\sigma \right) \right\|^2 
+ 2 \, \mathbb{E} \left\langle \nabla \mathcal{L}(\boldsymbol{\theta}_t),\ \frac{1}{B} \left( \sum_i \boldsymbol{z}_{t,i} + \boldsymbol{z}_\sigma \right) \right\rangle \\
&= \left\| \nabla \mathcal{L}(\boldsymbol{\theta}_t) \right\|^2 
+ \mathbb{E} \left\| \frac{1}{B} \sum_i \boldsymbol{z}_{t,i} \right\|^2 
+ \mathbb{E} \left\| \frac{1}{B} \boldsymbol{z}_\sigma \right\|^2 
+ \frac{2}{B} \sum_i \mathbb{E} \left\langle \boldsymbol{z}_{t,i},\ \boldsymbol{z}_\sigma \right\rangle.
\end{align*}

Also, we have
\[
\mathbb{E} \| \boldsymbol{z}_{t,i} \|^2 = \mathbb{E} \sum_I \boldsymbol{z}_{t,i}^2 = \sum_I \mathbb{E} \boldsymbol{z}_{t,i}^2 = \sum_I \boldsymbol{z}_{t,i}^2 = d \sigma^2,
\]
according to Equation~\ref{eq:mask_bound_alpha}, it follows that
\begin{equation}
\mathbb{E} \left\| \boldsymbol{m} \odot \left( \nabla \mathcal{L}(\boldsymbol{\theta}_t) + \frac{1}{B} \sum_I (\boldsymbol{z}_{t,i} + \boldsymbol{z}_\sigma) \right) \right\|^2 
\leq \alpha_t \left[\| \nabla \mathcal{L}(\boldsymbol{\theta}_t) \|^2 + \frac{\sigma_g^2}{B} + \frac{d \sigma^2}{B^2} \right]
\label{eq:masked_norm_upper_bound}
\end{equation}

Substituting Equation~\ref{eq:masked_norm_upper_bound} into Equation~\ref{eq:final_bound}, we obtain
\[
\eta \alpha_t \| \nabla \mathcal{L}(\boldsymbol{\theta}_t) \|^2 
\leq \mathbb{E}[\mathcal{L}(\boldsymbol{\theta}_t) - \mathcal{L}(\boldsymbol{\theta}_{t+1})] 
+ \frac{\alpha_t G \eta^2}{2} \left( \| \nabla \mathcal{L}(\boldsymbol{\theta}_t) \|^2 
+ \frac{\sigma_g^2}{B} + \frac{d \sigma^2}{B^2} \right),
\]

Taking $\eta = \frac{1}{G \sqrt{T}}$, we have
\[
\frac{\alpha_t}{G \sqrt{T}} \| \nabla \mathcal{L}(\boldsymbol{\theta}_t) \|^2 
\leq \mathbb{E}[\mathcal{L}(\boldsymbol{\theta}_t) - \mathcal{L}(\boldsymbol{\theta}_{t+1})] 
+ \frac{\alpha_t}{2 G T} \left( \| \nabla \mathcal{L}(\boldsymbol{\theta}_t) \|^2 
+ \frac{\sigma_g^2}{B} + \frac{d \sigma^2}{B^2} \right),
\]

We collect the coefficient of $\|\nabla \mathcal{L}(\boldsymbol{\theta}_t)\|^2$
\[
\left( \frac{1}{G\sqrt{T}} - \frac{1}{2GT} \right) \alpha_t \|\nabla \mathcal{L}(\boldsymbol{\theta}_t)\|^2 
\leq \mathbb{E}[\mathcal{L}(\boldsymbol{\theta}_t) - \mathcal{L}(\boldsymbol{\theta}_{t+1})] 
+ \frac{\alpha_t}{2GT} \left( \frac{\sigma_g^2}{B} + \frac{d\sigma^2}{B^2} \right),
\]
\[
\left( \frac{1}{G\sqrt{T}} - \frac{1}{2GT} \right) 
= \frac{2}{2G\sqrt{T}} - \frac{1}{2G\sqrt{T}} = \frac{2 - \frac{1}{\sqrt{T}}}{2G\sqrt{T}}.
\]

Since $T > 1$, we have $2 - \frac{1}{\sqrt{T}} > 1$,
therefore
\[
\left( \frac{1}{G\sqrt{T}} - \frac{1}{2GT} \right) 
= \frac{2 - \frac{1}{\sqrt{T}}}{2G\sqrt{T}} > \frac{1}{2G\sqrt{T}}.
\]

Thus, we obtain
\[
\frac{\alpha_t}{2G\sqrt{T}} \|\nabla \mathcal{L}(\boldsymbol{\theta}_t)\|^2 
\leq \mathbb{E}[\mathcal{L}(\boldsymbol{\theta}_t) - \mathcal{L}(\boldsymbol{\theta}_{t+1})] 
+ \frac{\alpha_t}{2GT} \left( \frac{\sigma_g^2}{B} + \frac{d\sigma^2}{B^2} \right).
\]

We have
\[
\| \nabla \mathcal{L}(\boldsymbol{\theta}_t) \|^2 
\leq \frac{2G \sqrt{T}}{\alpha_t} \, \mathbb{E}[\mathcal{L}(\boldsymbol{\theta}_t) - \mathcal{L}(\boldsymbol{\theta}_{t+1})] 
+ \frac{1}{\sqrt{T}} \left( \frac{\sigma_g^2}{B} + \frac{d \sigma^2}{B^2} \right).
\]

Taking average over $T$ steps
\[
\frac{1}{T} \sum_t \| \nabla \mathcal{L}(\boldsymbol{\theta}_t) \|^2 
\leq \frac{2G\sqrt{T}}{r} \cdot \frac{1}{T} \sum_t \mathbb{E}[\mathcal{L}(\boldsymbol{\theta}_t) - \mathcal{L}(\boldsymbol{\theta}_{t+1})] 
+ \frac{1}{\sqrt{T}} \left( \frac{\sigma_g^2}{B} + \frac{d \sigma^2}{B^2} \right),
\]

Since
\[
\mathbb{E}[\mathcal{L}(\boldsymbol{\theta}_{t+1})] \geq \min_\theta \mathcal{L}(\boldsymbol{\theta}),
\]
we have
\[
\mathbb{E}[\mathcal{L}(\boldsymbol{\theta}_1) - \mathcal{L}(\boldsymbol{\theta}_{t+1})] 
= \mathbb{E}[\mathcal{L}(\boldsymbol{\theta}_1)] - \mathbb{E}[\mathcal{L}(\boldsymbol{\theta}_{t+1})] 
\leq \mathbb{E}[\mathcal{L}(\boldsymbol{\theta}_1)] - \min_\theta \mathcal{L}(\boldsymbol{\theta}).
\]

Therefore,
\[
\frac{1}{T} \sum_t \| \nabla \mathcal{L}(\boldsymbol{\theta}_t) \|^2 
\leq \frac{2G}{\alpha_t \sqrt{T}} \left( \mathbb{E}[\mathcal{L}(\boldsymbol{\theta}_1)] - \min_\theta \mathcal{L}(\boldsymbol{\theta}) \right) 
+ \frac{1}{\sqrt{T}} \left( \frac{\sigma_g^2}{B} + \frac{d \sigma^2}{B^2} \right).
\]

\end{proof}

\subsection{Proof of Theorem~\ref{the:topk+clip+DPSGD}}\label{proof:topk+clip+DPSGD}

\begin{proof}
This proof aims to upper bound the expected alignment between the true gradient and the privatized update $\tilde{\boldsymbol{g}}_t$ under coordinate-wise masking and additive Gaussian noise. Starting from the $G$-smoothness of the loss function, we derive a one-step descent inequality by substituting the noisy SGD update rule. The inner product $\langle \nabla \mathcal{L}, \tilde{\boldsymbol{g}}_t \rangle$ is then isolated and upper bounded by the function value decrease and the second-order moment of the update. The masked noise contribution is controlled via a sparsity-aware energy retention factor $\alpha_t$, and the variance of the Gaussian noise is explicitly computed. Combining these steps yields a convergence-style upper bound that reveals how $\alpha_t$, $B$, $d$, and $\sigma$ jointly affect training dynamics under differential privacy.
By the $G$-smoothness of $\mathcal{L}$, we have
\[
\mathcal{L}(\boldsymbol{\theta}_{t+1}) 
\leq \mathcal{L}(\boldsymbol{\theta}_t) + \langle \nabla \mathcal{L}(\boldsymbol{\theta}_t), (\boldsymbol{\theta}_{t+1} - \boldsymbol{\theta}_t) \rangle + \frac{G}{2} \| \boldsymbol{\theta}_{t+1} - \boldsymbol{\theta}_t \|^2,
\]
using the update rule
\[
\boldsymbol{\theta}_{t+1} = \boldsymbol{\theta}_t - \eta \left( \tilde{\boldsymbol{g}}_t + \boldsymbol{m} \odot \boldsymbol{z}_{DP} \right),
\]
where the average clipped gradient $\tilde{\boldsymbol{g}}_t$ is defined by
\[
\tilde{\boldsymbol{g}}_t = \frac{1}{B} \sum_i \frac{\boldsymbol{g}_t'(\boldsymbol{x}_i)}{\max(1, C / \|\boldsymbol{g}_t'(\boldsymbol{x}_i)\|)},
\]
we get
\begin{align*}
\mathcal{L}(\boldsymbol{\theta}_{t+1}) 
&= \mathcal{L}(\boldsymbol{\theta}_t) 
- \eta \langle \nabla \mathcal{L}(\boldsymbol{\theta}_t), \tilde{\boldsymbol{g}}_t + \boldsymbol{m} \odot \boldsymbol{z}_{DP} \rangle 
+ \frac{G \eta^2}{2} \left\| \tilde{\boldsymbol{g}}_t + \boldsymbol{m} \odot \boldsymbol{z}_{DP} \right\|^2 \\
&= \mathcal{L}(\boldsymbol{\theta}_t) 
- \eta \langle \nabla \mathcal{L}(\boldsymbol{\theta}_t), \tilde{\boldsymbol{g}}_t \rangle 
- \eta \langle \nabla \mathcal{L}(\boldsymbol{\theta}_t), \boldsymbol{m} \odot \boldsymbol{z}_{DP} \rangle 
+ \frac{G \eta^2}{2} \left\| \tilde{\boldsymbol{g}}_t + \boldsymbol{m} \odot \boldsymbol{z}_{DP} \right\|^2,
\end{align*}
We take expectation and rearrange
\begin{align*}
\mathbb{E}[\langle \nabla \mathcal{L}(\boldsymbol{\theta}_t), \tilde{\boldsymbol{g}}_t \rangle] 
&= \frac{1}{\eta} \mathbb{E}[\mathcal{L}(\boldsymbol{\theta}_t) - \mathcal{L}(\boldsymbol{\theta}_{t+1})] 
- \mathbb{E}[\langle \nabla \mathcal{L}(\boldsymbol{\theta}_t), \boldsymbol{m} \odot \boldsymbol{z}_{DP}] + \frac{G \eta}{2} \mathbb{E}[\| \tilde{\boldsymbol{g}}_t + \boldsymbol{m} \odot \boldsymbol{z}_{DP} \|^2] \\
\\
&= \frac{1}{\eta} \mathbb{E}[\mathcal{L}(\boldsymbol{\theta}_t) - \mathcal{L}(\boldsymbol{\theta}_{t+1})] 
+ \frac{G \eta}{2} \left( \mathbb{E}[\| \tilde{\boldsymbol{g}}_t \|^2] + \mathbb{E}[\| \boldsymbol{m} \odot \boldsymbol{z}_{DP} \|^2] \right).
\end{align*}

Since (by Corollary~\ref{cor:topk_mask_expectation})
\[
\mathbb{E} \left\| \boldsymbol{m}^{(k)} \odot \boldsymbol{v} \right\|^2 
\leq \alpha_t \cdot \mathbb{E} \|\boldsymbol{v}\|^2,
\]
we have
\[
\mathbb{E}[\| \boldsymbol{m} \odot \boldsymbol{z}_{DP} \|^2] 
\leq \alpha_t \cdot \mathbb{E}[\|\boldsymbol{z}_{DP}\|^2].
\]

Therefore
\[
\mathbb{E}[\langle \nabla \mathcal{L}(\boldsymbol{\theta}_t), \tilde{\boldsymbol{g}}_t \rangle] 
\leq \frac{1}{\eta} \mathbb{E}[\mathcal{L}(\boldsymbol{\theta}_t) - \mathcal{L}(\boldsymbol{\theta}_{t+1})] 
+ \frac{G \eta}{2} \left( \mathbb{E}[\| \tilde{\boldsymbol{g}}_t \|^2] + \alpha_t \cdot \mathbb{E}[\| \boldsymbol{z}_{DP} \|^2] \right).
\]

Since each $\boldsymbol{z}_i \sim \mathcal{N}(\boldsymbol{0}, \sigma^2 C^2 \boldsymbol{I}_d)$ is independent, the average noise vector is
\[
\boldsymbol{z}_{DP} := \frac{1}{B} \sum_{i=1}^B \boldsymbol{z}_i \sim \mathcal{N}(\boldsymbol{0}, \frac{\sigma^2 C^2}{B} \boldsymbol{I}_d),
\]
and thus
\[
\mathbb{E}[\|\boldsymbol{z}_{DP}\|^2] = \operatorname{tr}(\operatorname{Cov}(\boldsymbol{z}_{DP})) = d \cdot \frac{\sigma^2 C^2}{B}.
\]

Therefore
\begin{align*}
\mathbb{E}[\langle \nabla \mathcal{L}(\boldsymbol{\theta}_t), \tilde{\boldsymbol{g}}_t \rangle]
&\leq \frac{1}{\eta} \mathbb{E}[\mathcal{L}(\boldsymbol{\theta}_t) - \mathcal{L}(\boldsymbol{\theta}_{t+1})]
- 0 + \frac{G \eta}{2} \left( \mathbb{E}[\| \tilde{\boldsymbol{g}}_t \|^2] + \alpha_t \cdot \frac{d \sigma^2 C^2}{B} \right) \nonumber \\
&\leq \frac{1}{\eta} \mathbb{E}[\mathcal{L}(\boldsymbol{\theta}_t) - \mathcal{L}(\boldsymbol{\theta}_{t+1})]
+ \frac{G \eta}{2} \left( C^2 + \alpha_t \cdot \frac{d \sigma^2 C^2}{B} \right).
\end{align*}

Thus,
\[
\mathbb{E}[\langle \nabla \mathcal{L}(\boldsymbol{\theta}_t), \tilde{\boldsymbol{g}}_t \rangle]
\leq \frac{1}{\eta} \mathbb{E}[\mathcal{L}(\boldsymbol{\theta}_t) - \mathcal{L}(\boldsymbol{\theta}_{t+1})]
+ \frac{G \eta}{2} \left( C^2 + \alpha_t \cdot \frac{d \sigma^2 C^2}{B} \right).
\]
\end{proof}

\vskip 0.2in
\bibliography{sample}

\end{document}